\RequirePackage{fix-cm}
\documentclass[twocolumn]{svjour3}          
\smartqed  
\usepackage{graphicx}
\usepackage{amsmath}
\usepackage{amssymb}
\usepackage{booktabs}
\usepackage{algpseudocode}
\usepackage{algorithm}
\usepackage{mathtools}
\usepackage{color}
\usepackage{multirow,graphicx,paralist}
\usepackage{ colortbl}
\usepackage{ tabularx}
\usepackage{pgfplots}
\usepackage{enumitem}
\pgfplotsset{compat=1.3}
\usepackage{multicol}
\usepackage{epstopdf}

\definecolor{Gray}{gray}{0.85}
\definecolor{LightCyan}{rgb}{0.88,1,1}
\definecolor{antiquefuchsia}{rgb}{0.57, 0.36, 0.51}
\definecolor{bleudefrance}{rgb}{0.19, 0.55, 0.91}
\usepackage[export]{adjustbox}
\usepackage[font=small,labelfont=bf]{caption}
\definecolor{maroon}{cmyk}{0,0.87,0.68,0.32}
\newcommand{\blue}[1]{\textcolor{black}{#1}}

\def\infinity{\rotatebox{90}{8}}

\usepackage{upquote}

\usepackage[letterpaper=true,colorlinks=true, linkbordercolor =red, urlcolor=black, bookmarks=false]{hyperref}
\begin{document} \sloppy

\title{SegMix: Co-occurrence Driven Mixup for Semantic Segmentation and Adversarial Robustness 
}


\author{Md Amirul Islam         \and
        Matthew Kowal \and Konstantinos G. Derpanis  \and Neil D. B. Bruce 
}


\institute{Md Amirul Islam \at
              Ryerson University, Vector Institute for AI \\
              \email{amirul@cs.ryerson.ca}           
           \and
           Matthew Kowal \at
              York University, Vector Institute for AI \\
              \email{m2kowal@eecs.york.ca}  
            \and
            Konstantinos G. Derpanis \at
              York University, Samsung AI Research Center Toronto, Vector Institute for AI \\
              \email{kosta@eecs.york.ca}  
                   \and
            Neil D. B. Bruce \at
              University of Guelph, Vector Institute for AI \\
              \email{brucen@uoguelph.ca}  
}


\maketitle

\begin{abstract}
In this paper, we present a strategy for training convolutional neural networks to effectively resolve interference arising from competing hypotheses relating to inter-categorical information throughout the network. The premise is based on the notion of feature binding, which is defined as the process by which activations spread across space and layers in the network are successfully integrated to arrive at a correct inference decision. \blue{In our work, this is accomplished for the task of dense image labelling by blending images based on (i) categorical clustering or (ii)  the co-occurrence likelihood of categories. We then train a \textit{feature binding} network which simultaneously segments and separates the blended images}. Subsequent feature denoising to suppress noisy activations reveals additional desirable properties and high degrees of successful predictions. Through this process, we reveal a general mechanism, distinct from any prior methods, for boosting the performance of the base segmentation and saliency network while simultaneously increasing robustness to adversarial attacks. 
\keywords{Categorical Mixup \and Feature Binding \and Semantic Segmentation\and Adversarial Robustness}
\end{abstract}

\section{Introduction}\label{sec:intro}

\begin{figure}[t ]
	\begin{center}
		\includegraphics[width=0.49\textwidth]{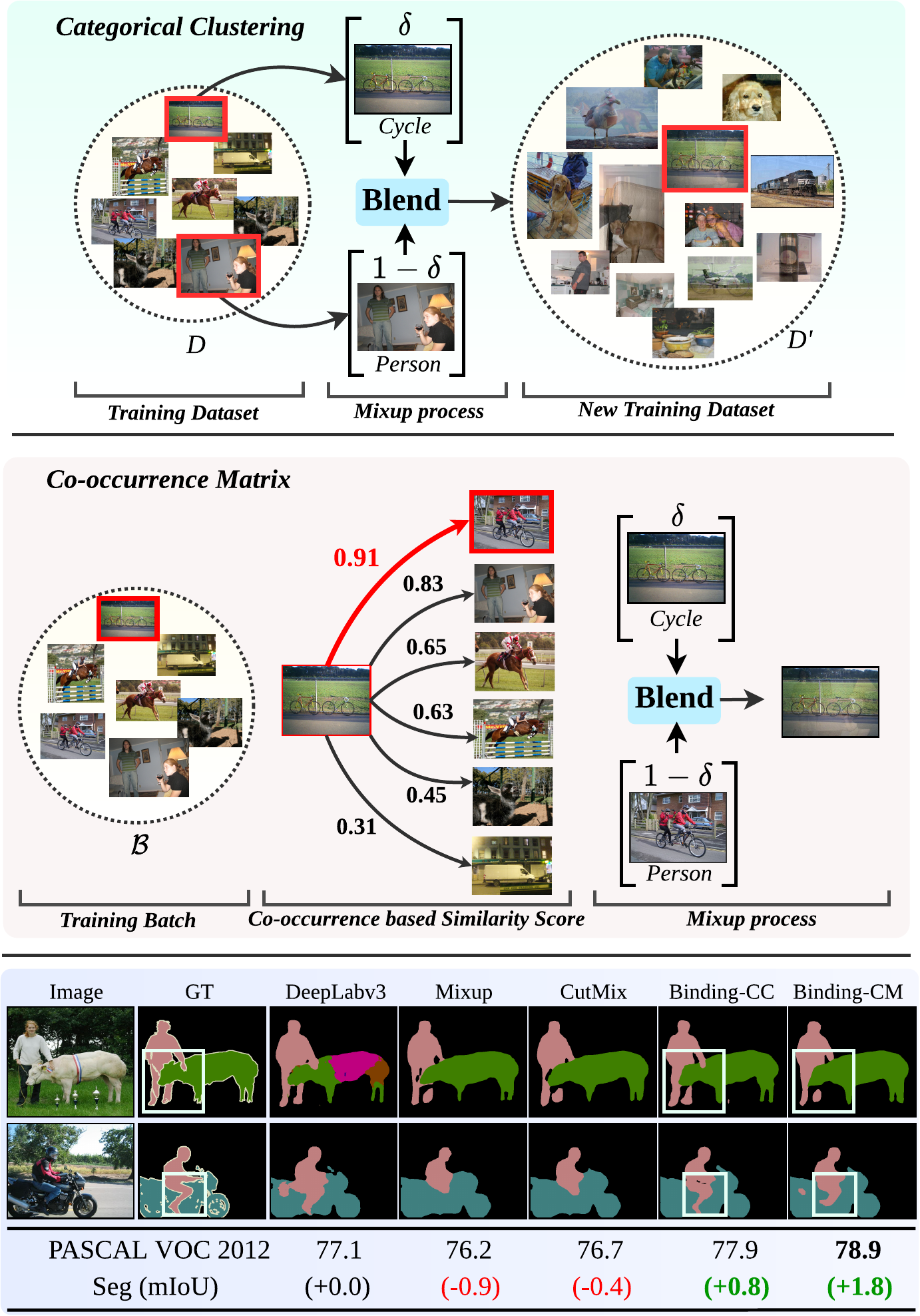}
		\caption{\textbf{Top}: Overview of our category-specific image blending to create a new source dataset ($D'$). A segmentation network is trained with $D'$ to simultaneously separate and segment both source and target images. \textbf{Middle}: Overview of our co-occurrence based image blending to create a mixed sample in the training batch, while training to simultaneously separate and segment both source and target images. \textbf{Bottom}: Results of our \textit{feature binding} methods (Binding-CC and Binding-CM), Mixup~\cite{zhang2017mixup}, and CutMix~\cite{yun2019cutmix} on the PASCAL VOC 2012~\cite{everingham2015pascal} segmentation task. Note that our methods significantly improve the overall performance.}
		\label{fig:intro}
	\end{center}
\end{figure}
The advent of Deep Neural Networks (DNNs) has seen overwhelming improvement in dense image labeling tasks~\cite{long15_cvpr,noh15_iccv,badrinarayanan15_arxiv,ghiasi2016laplacian,zhao2017pyramid,Islam_2017_CVPR,chen2018deeplab,cvpr18_rank,li2016iterative,islam2017label,islam2018gated,karim2019recurrent,he2017mask,karim2020distributed}, however, for some common benchmarks~\cite{everingham2015pascal} the rate of improvement has slowed down. While one might assume that barriers to further improvement require changes at the architectural level, it has also been borne out that pre-training across a variety of datasets~\cite{russakovsky2015imagenet,lin2014microsoft} can improve performance that exceeds improvements seen from changing the model architecture. However, there are challenging scenarios for which DNNs have difficulty regardless of pre-training or architectural changes, such as highly occluded scenes, or objects appearing out of their normal context~\cite{singh2020don}. It is not clear though, for dense image labeling tasks, how to resolve these specific scenarios for more robust prediction quality on a per-pixel level.
In particular, one might expect that failures in correctly predicting labels for an image are more likely to be seen for challenging cases once a critical performance threshold has been reached.\\

A question that naturally follows from this line of reasoning is: How can the number of locally challenging cases be increased, or the problem made more difficult in general? In this paper, we address this problem using a principled approach to improve performance and that also implies a more general form of robustness. As inspiration, we look to a paradigm discussed often in the realm of human vision: the binding problem~\cite{treisman1998feature,shipp2009feature}. The crux of this problem is that given a complex decomposition of an image into features that represent different concepts, or different parts of the image, how does one proceed to successfully relate activations corresponding to common sources in the input image to label a \emph{whole} from its parts, or separate objects. Motivated by the binding problem, a successful solution in the computer vision domain should rely on both determining correspondences in activations among features that represent disparate concepts, and also to associate activations tied to related features that are subject to spatial separation in the image. To address similar issues for the image classification task, recent studies~\cite{tokozume2018between,zhang2017mixup,yun2019cutmix,kim2020puzzle,hendrycks2019augmix,verma2019manifold,chang2020mixup} have considered mixing two image examples with constraints on the distribution of features. However, these methods suffer from biases in the dataset~\cite{islam2021shape} used, as they have no strategy when deciding on which images to mix which is crucial for the dense labeling problem. Additionally, these strategies do not adequately separate information from different sources in the image as they only require the network to make a single (classification) prediction during training.

In our work, the means of solving the feature binding problem takes a direct form, which involves training networks on specially designed training data of mixed images (see Fig.~\ref{fig:intro} (top and middle)) to simultaneously address problems of dense image labeling~\cite{long15_cvpr,chen15_iclr,noh15_iccv}, and blind source separation~\cite{georgiev2005sparse,huang2015joint}. Humans show a surprising level of capability in interpreting a superposition (e.g., average) of two images, both interpreting the contents of each scene and determining the membership of local patterns within a given scene. The underlying premise of this work involves producing networks capable of simultaneously performing dense image labeling for pairs of images while also separating labels according to the source images. If one selects pairs on the basis of a weighted average, this allows treatment of the corresponding dense image labeling problem in the absence of source separation by extension. This process supports several objectives: (i) it significantly increases the number of occurrences that are locally ambiguous that need to be resolved to produce a correct categorical assignment, (ii) it forces broader spatial context to be considered in making categorical assignments, and (iii) it stands to create more powerful networks for standard dense labeling tasks and dealing with adversarial perturbations by forcing explicit requirements on how the network uses the input. The end goal of our procedure is to improve overall performance as well as increase the prediction quality on complex images (see Fig.~\ref{fig:intro} (bottom)), heavily occluded scenes, and also invoke robustness to challenging adversarial inputs. \\

\blue{The contribution of this paper extends from the approach presented in our prior work~\cite{islam2020feature} which introduced a categorical clustering based mixup strategy to generate a new training dataset. In addition, we also proposed a binding network~\cite{islam2020feature} to simultaneously perform dense image labeling for pairs of images while also separating labels according to the source image classes. We extend our prior work in the following respects:}

\begin{itemize}
  \setlength{\itemsep}{5.3pt}

    \item \blue{We introduce a new and efficient \textit{co-occurrence matrix} based \textit{mixup} strategy which exploits co-occurrence likelihood of semantic categories from the dataset in the mixup process. This technique trains the network to separate semantic objects in commonly occurring complex scenes with high degrees of occlusion.}
    
    \item \blue{We show, through extensive quantitative and qualitative experiments, that our newly introduced mixup technique outperforms our previous categorical clustering based technique~\cite{islam2020feature} and recent mixing methods~\cite{zhang2017mixup,yun2019cutmix} on the PASCAL VOC 2012 dataset~\cite{everingham2015pascal}, while simultaneously being less computationally expensive and maintaining robustness to adversarial attacks.}
    
    \item \blue{We evaluate our newly introduced technique for an additional task, salient object detection, which shows improvements over the baselines.}
    
    \item \blue{We provide an in-depth analysis and ablations of the introduced co-occurrence based mixup technique to  show its influence in improving performance and robustness.}

\end{itemize}

\blue{The paper is structured as follows: we discuss related work in Sec.~\ref{sec:related}. In Sec.~\ref{sec:approach}, we first introduce two different image mixup techniques for the task of semantic segmentation followed by the feature binding network for simultaneous dense labeling and source separation. Subsequently, we discuss the training procedure, and present the experimental results in Sec.~\ref{sec:exp}. Finally, we provide extensive ablation studies in Sec.~\ref{sec:ablation}.}

\section{Related Work}\label{sec:related}
\noindent \textbf{Semantic Segmentation.} Existing Convolutional Neural Network (CNN) based works~\cite{long15_cvpr,chen15_iclr,noh15_iccv,badrinarayanan15_arxiv,ghiasi2016laplacian,chen2017rethinking,takikawa2019gated,islam2020feature} have shown widespread success on dense image prediction tasks (e.g., semantic segmentation). The feature representations produced in the top layers of shallower and deeper CNNs~\cite{krizhevsky12_nips,simonyan15_iclr,szegedy2015going,he2016deep} carry a strong semantic representation, perhaps at the expense of retaining spatial details required for dense prediction due to the poor spatial resolution among deeper layers. 
In particular, atrous convolution~\cite{chen2017rethinking,yu2015multi,chen2018deeplab}, encoder-decoder structures~\cite{badrinarayanan15_arxiv,noh15_iccv} and pyramid pooling~\cite{zhao2017pyramid,chen2018deeplab} have been employed to decode low resolution feature maps, increase the contextual view, and capture context at different ranges of spatial precision, respectively. \\

\noindent \textbf{Data Augmentation.} \blue{Existing methods~\cite{bishop1995training,krizhevsky12_nips,hendrycks2019augmix,kim2020puzzle} introduced data augmentation based techniques to regularize the training of CNNs. These techniques regularize the models from over-fitting to the training distribution (e.g., categorical biases) and also improve the generalization ability by generating extra training samples given the original training set. Most commonly used data augmentation strategies are random cropping, horizontal flipping~\cite{krizhevsky12_nips}, and adding random noise~\cite{bishop1995training}. Recently proposed data augmentation techniques, termed AugMix~\cite{hendrycks2019augmix} and PuzzleMix~\cite{kim2020puzzle}, were designed to improve the generalization performance and robustness against corruptions. However, these techniques are extensively evaluated for image classification and its unclear if these techniques will perform better for dense labeling tasks. In contrast, our proposed approach can be complementary to these techniques and could be applied in conjunction to further improve the dense labeling performance and robustness. }  \\

\noindent \textbf{Mixup-based Augmentation.} \blue{More closely related to our work, contributions~\cite{yun2019cutmix,zhang2017mixup,tokozume2018between,inoue2018data,cubuk2019autoaugment,french2019semi,harris2020fmix,chou2020remix} on data augmentation based techniques share a similar idea of mixing two randomly selected samples to create new training data for the image classification or localization task. Between-Class (BC) learning~\cite{tokozume2018between} showed that randomly mixing training samples can lead to better separation between categories based on the feature distribution. Mixup~\cite{zhang2017mixup} shares a similar idea of training a network by mixing the data that regularizes the network and increases the robustness against adversarial examples. Manifold Mixup~\cite{verma2019manifold} extends Mixup~\cite{zhang2017mixup} from input space to feature space and showed improvement on overall performance. Further, Guo et al.~\cite{guo2019mixup} proposed an adaptive Mixup technique to prevent the generation of improper mixed data. CutMix~\cite{yun2019cutmix} further proposed to overlay a cropped area of an input image to another. However, these methods randomly blend images and may generate non-optimal training samples according to object distributions, which might be problematic for more complex dense labeling tasks. Our proposed techniques aim to address this issue by utilizing category-level information in the mixup process. Our proposed framework differs from the above existing works in that: (i) the network performs simultaneous dense prediction and source separation to achieve superior dense labeling and adversarial robustness; whereas, other techniques are focused mainly on image classification or object localization while using a single output, (ii) previous methods either mix labels as the ground truth or use the label from only one sample, while we use both ground truth labels independently, and (iii) samples are chosen randomly for Mixup~\cite{zhang2017mixup} and CutMix~\cite{yun2019cutmix} while we use two intuitive strategies (categorical clustering, Sec.~\ref{sec:cc} and Co-occurrence matrix, Sec.~\ref{sec:cm}).} \\

\blue{The groundwork for some of what is presented in this paper appeared previously~\cite{islam2020feature}, in which we introduced a categorical clustering based mixup strategy to generate a new training dataset followed by training a network for source separation with the ultimate goal of semantic segmentation. However, due to the size of the training dataset, the computational load during training was significant. In this work, we address this training inefficiency issue by introducing an intuitive mixup technique which considers the co-occurrence likelihood of semantic categories before mixing two images. The main advantage of this new technique is the training sample generation is done online within a batch instead of creating a new large dataset offline. We provide an in-depth analysis of the newly introduced mixup technique to show its influence in improving performance and robustness.}

\section{Proposed Method}\label{sec:approach}
In the broader context of investigating approaches motivated by the feature binding problem, we propose a novel framework capable of solving the dense labeling problem. Our proposed framework consists of three key steps: (i) we first apply a blending technique (Sec.~\ref{sec:fbt}) on the training dataset either offline (Sec.~\ref{sec:cc}) or online (Sec.~\ref{sec:cm}), (ii) we train a CNN using the generated data that simultaneously produces dense predictions and source separations (Sec.~\ref{sec:FBNet}), and (iii) we denoise the learned features from the feature binding process by fine-tuning on standard data (Sec.~\ref{sec:denoise}). 

\subsection{Category-Dependent Image Blending}\label{sec:fbt}
Recent works~\cite{zhang2017mixup,yun2019cutmix,tokozume2018between,inoue2018data,cubuk2019autoaugment} simply mix two randomly selected samples to create new training data for image classification or object localization. Exploring a similar direction, we are interested in solving dense prediction tasks (e.g., semantic segmentation, salient object detection) in a way that provides separation based on mixed source images. The traditional way~\cite{zhang2017mixup,tokozume2018between,inoue2018data} of combining two images is by a weighted average which implies that the contents of both scenes appear with varying contrast. Randomly combining two source images to achieve the desired objective is a more significant challenge than one might expect in the context of dense prediction. One challenge is the categorical bias of the dataset (e.g., mostly the \textit{person} images will be combined with all other categories, since \textit{person} is the most common category in PASCAL VOC 2012) across the newly generated training set. Previous methods~\cite{zhang2017mixup,tokozume2018between,inoue2018data}, randomly select images to combine, results in a new data distribution with similar inherent biases as the original dataset. 

\blue{To overcome these limitations, we introduce two different image blending techniques to create new training data, denoted as \textit{categorical clustering} and \textit{co-occurrence matrix}. To blend based on categorical clustering, we augment the PASCAL VOC 2012~\cite{everingham2015pascal} training dataset based on categorical clustering to generate a new training set in a form that accounts for source separation and dense prediction. Categorical clustering combines images based on a uniform distribution across categories. For the co-occurrence matrix-based strategy, we consider the co-occurrence likelihood between semantic objects in the blending process to generate new training data. The main difference between these two blending techniques is the way that new training data is generated. The former one generates a new training dataset offline while the latter one blends images in the training batch.}
Thorough experimentation with our proposed mixing strategies show improvements in the network's ability to separate competing categorical features and can generalize these improvements to various challenging scenarios, such as segmenting out-of-context objects or highly occluded scenes.

\subsubsection{Categorical Clustering}\label{sec:cc} We first generate 20 different clusters of images, where each cluster contains images of a certain category from VOC 2012. For each training sample in a cluster, we linearly combine it with a random sample from each of the 19 other clusters. For example, given a training sample $\mathcal{I}_{a}$ from the \textit{person} cluster we randomly choose a sample $\mathcal{I}_{b}$ from another categorical cluster and combine them to obtain a new sample, $\mathcal{I}_{fb}$: 
\begin{equation}\label{eq:combine}
\begin{split}
   \mathcal{I}_{fb} = \delta \ast \mathcal{I}_{a} + (1-\delta) \ast \mathcal{I}_{b} ,
   \end{split}
\end{equation}
where $\delta$ denotes the randomly chosen weight that is applied to each image. We assign the weight such that the source image ($\mathcal{I}_{a}$) has more weight compared to the random one ($\mathcal{I}_{b}$). In our experiments, we sample $\delta$ uniformly from a range of $[0.7-1]$ for each image pair. Note that for one sample (e.g., the \textit{person} cluster), we generate 19 new samples. We continue to generate new training samples for the other remaining images in the person cluster and perform the same operation for images in other clusters.

\begin{algorithm*}
		\caption{\textbf{Co-occurrence based Image Blending} }\label{cooccur}
		\begin{algorithmic}[1]
			
			\Statex \textbf{Input}: Training Batch $\mathcal{B}=\{I, G\}$; Co-occurrence matrix $\mathcal{C}_{fb}$, $\alpha$, max unique Category threshold, $\gamma$
			\Statex \textbf{Output}: New training batch $\mathcal{B}^\prime$
			
			\vspace{0.2cm}
			\If {$\alpha >$ 0}
				\State $\delta$  $\leftarrow$ random.beta($\alpha$, $\alpha$) \Comment{Generate $\delta$ from beta distribution if $\alpha > 0$}

			\Else
                \State $\delta$  $\leftarrow$  1			
			\EndIf
				\vspace{0.2cm}
			\State $\mathcal{B}^\prime$ $\leftarrow$ \{\}, \hspace{0.2cm} $\mathcal{Y}_1^\prime$$\leftarrow$ \{\}, \hspace{0.2cm} $\mathcal{Y}_2^\prime$$\leftarrow$ \{\}
		
				\vspace{0.2cm}
			\For{ $\mathcal{I}^k$ $\in$ $\mathcal{B}$  }
				\vspace{0.2cm}
			\State similarity-score $\leftarrow$ zeros(len($\mathcal{B}$)) \Comment{Store similarity score with each sample in the batch other than $\mathcal{I}^k$}
			
			\State mixed-category-list $\leftarrow$ zeros(len($\mathcal{B}$)) \Comment{Store total number of unique semantic categories}
				\vspace{0.2cm}
			
			\For{$\mathcal{I}^m$ $\in$ $\mathcal{B}$}
				\vspace{0.2cm}
			\If {$k \neq m$}
			
			\Statex \hspace{1cm}\textcolor{cyan}{/* \textit{Compute the unique semantic categories}  \hspace{0.1cm}    */}
			\State \textbf{K-unique} $\leftarrow$ unique(Ground-truth($\mathcal{I}^k$))
			\State \textbf{M-unique} $\leftarrow$ unique(Ground-truth($\mathcal{I}^m$))
			
			\State Remove the background class index and the ignore class index from both K-unique and M-unique
			
				\vspace{0.2cm}
			\Statex \hspace{1cm}\textcolor{cyan}{/* \textit{Consider pairs with total categories $\leq$ 2}  \hspace{0.1cm}    */}
			\If {len(K-unique) $\geq$ 1 \& len(M-unique) $\geq$ 1  }
            \Statex \hspace{1cm}\textcolor{cyan}{/* \textit{Initialize co-occurrence score to 0}  \hspace{0.1cm}    */}
            \State cooccurrence-score $\leftarrow$ 0
            \Statex \hspace{0.9cm}\textcolor{cyan}{/* \textit{Compute total co-occurrence score}  \hspace{0.1cm}    */}
			\For{  i $\leftarrow$ 1 to len(M-unique)} 
              	\For{  j $\leftarrow$ 1 to len(K-unique)} 
                    \State  cooccurrence-score $\leftarrow$ cooccurrence-score +  $\mathcal{C}_{fb}$[M-unique[i]][K-unique[j]]
              	\EndFor
            \EndFor
            \vspace{0.2cm}
            \State similarity-score [m] $\leftarrow$ cooccurrence-score
            \State mixed-category-list [m] $\leftarrow$ len(M-inique) + len(K-unique)
			\EndIf
			
			\EndIf
 			\EndFor
 				\vspace{0.2cm}
		
			 \State  top-sim-idx $\leftarrow$ \texttt{argmax}(similarity-score)
			 \Comment{ Choose the index with highest similarity score}
			 \vspace{0.2cm}
			 \If {mixed-category-list[top-sim-idx] $>$ $\gamma$}
			\State  $\delta$ $\leftarrow$ 0.9
			\EndIf

			\vspace{0.2cm}
			\State $\mathcal{I}^k_{fb}\leftarrow$$\delta \ast \mathcal{I}^k+$ $(1-\delta) \ast \mathcal{I}^{top-sim-idx}$  
			\Comment{Mix $\mathcal{I}^k$ with sample with highest similarity score}
			\State $\mathcal{Y}_1^\prime$ $\leftarrow$ $\mathcal{Y}[k]$
			, \hspace{0.2cm} $\mathcal{Y}_2^\prime$ $\leftarrow$ $\mathcal{Y}[top-sim-idx]$ \Comment{Choose the corresponding ground-truth segmentation map}
			\vspace{0.2cm}
			
			\State $ \mathcal{B}^\prime$ $\leftarrow$ $\{ \mathcal{I}^k_{fb}, \mathcal{Y}_1^\prime, \mathcal{Y}_2^\prime\} $ \Comment{Mixed training sample in Batch, $\mathcal{B}^\prime$}

			\EndFor
		\end{algorithmic}
		\label{alg:cooccur}
	\end{algorithm*}

\subsubsection{Co-occurrence Matrix}\label{sec:cm} \blue{We propose an additional technique that blends images based on the co-occurrences among semantic categories (i.e., probability of appearing together in an image). Towards this goal, we first calculate the co-occurrence matrix, $\mathcal{C}_{fb}\in{\mathbb{R}^{n\times n}}$ (\textit{n}= number of categories) using Eq.~\ref{eq:cocur} that contains the number of times two semantic categories co-occur within the training set. For computing co-occurrence score between category, $i$ and $j$, we formalize the equation as follows:}

\begin{align} \label{eq:cocur}
\mathcal{C}_{fb}(i,j) = \sum_{x=1}^{X}\sum_{y=1}^{Y}   \left\{ \begin{array}{cc}
1 & \hspace{1cm} \texttt{if}\hspace{1.9mm} \mathcal{I}_s(x,y)=i \hspace{1.5mm} \& \hspace{1.5mm} c=j \\
0 & \hspace{1cm} \texttt{otherwise}
\end{array}
\right. 
\end{align}
where $i$ and $j$ are the object classes (i.e., pixel values in ground-truth); $x$ and $y$ are the spatial position in the image $\mathcal{I}_s$.\\

Algorithm~\ref{alg:cooccur} describes the set of steps for generating new training samples in a batch based on the pre-computed co-occurrence matrix.

\begin{figure*} [t]
	\begin{center}
		\includegraphics[width=0.98\textwidth]{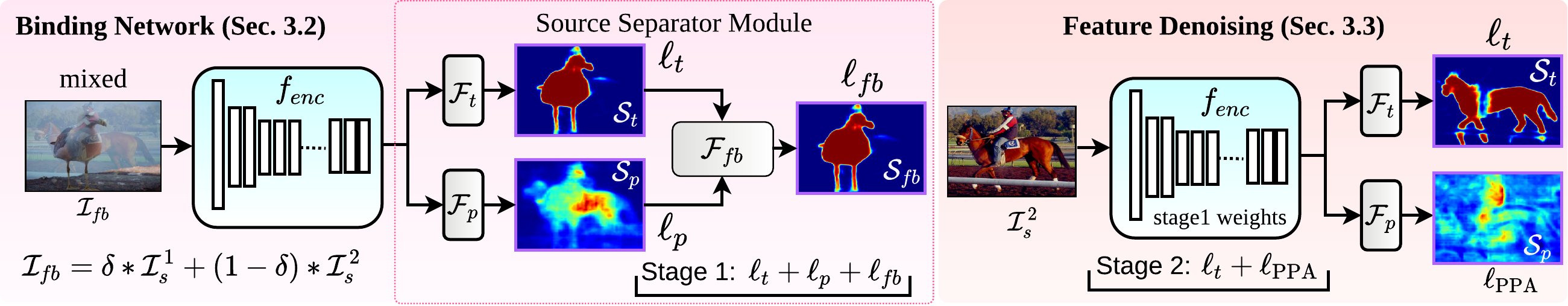}
		\caption{\blue{An illustration of our proposed framework. At the data end, categorical collisions are created with a \textit{dominant} ($\mathcal{I}_a$) and \textit{phantom} ($\mathcal{I}_{b}$) image. \textbf{Stage 1:} The network is trained on mixed data ($\mathcal{I}_{fb}$) to perform simultaneous dense labeling and source separation. We use the labels of both \textit{dominant} and \textit{phantom} images as the targets for two separate output channels. \textbf{Stage 2:} Fine-tuning on standard data to further promote desirable properties along the two dimensions of base performance and robustness to perturbations. In this stage, the \textit{phantom} activation of the second channel is suppressed. Confidence maps are plotted with the `Jet' colormap, where red and blue indicates higher and lower confidence, respectively.}}
		\label{fig:architecture}
	\end{center}
\end{figure*}

\blue{In summary: for each training sample, $I_a$ in a batch, $B$, of size $N$, we compute a similarity score with the other $N-1$ samples based on the pre-computed co-occurrence matrix. We pick the sample with highest similarity score, $I_b$, to be combined with the sample $I_a$. Finally, we apply Eq.~\ref{eq:combine} to generate a new blended training sample. Similar to the clustering based blending technique, we randomly choose $\delta$ and assign more weight on the source image, $\mathcal{I}_{a}$ compared to the target image, $\mathcal{I}_{b}$.
The intuition of assigning higher weight on the source image is that the semantic segmentation task requires to learn the context of the semantic objects for accurate per-pixel labeling. However, blending two images with a large number of semantic categories with lower mixed ratio (i.e., assigning more weight on the \textit{target} image) significantly increases the possibility of destroying the contextual information as well as introducing unlikely samples into the training set when considering the object distribution. For example, PASCAL VOC has very few images with 7+ objects in it, and therefore training on blended images with this many objects may add noise during training when more weight is assigned to the target image. Therefore, we choose a threshold for the maximum number of unique semantic categories, $\gamma$. If the total number of unique semantic categories in the chosen pair is greater than a certain threshold, we set the mixing ratio, $\delta$ to 0.9. 
} \\

\blue{While there exist alternatives~\cite{zhang2017mixup,devries2017improved,yun2019cutmix} for combining pairs of images to generate a training set suitable for source separation training, our intuitive methods are simple to implement and achieve strong performance on a variety of metrics (see Sec.~\ref{sec:exp}). Exploring further methods to combine and augment the training set is an interesting and nuanced problem to be studied further in the context of dense image labeling.}



\subsection{Feature Binding Network}\label{sec:FBNet}
In this section, we present a fully convolutional feature binding network in the context of dense prediction. Figure~\ref{fig:architecture} illustrates the overall pipeline of our proposed method. \\

\noindent \textbf{Overview and Notations.} During training, our goal is to produce a dense prediction of a source image, $\mathcal{I}_{fb}$, given a pair of images ($\text{\textit{dominant}: } \mathcal{I}_{a}, \text{\textit{phantom}: } \mathcal{I}_{b}$) and their corresponding ground-truth, ($\mathcal{G}_{a}, \mathcal{G}_{b}$). Note that each source image, $\mathcal{I}_{fb}$ in the new set is a weighted combination of two images ($\mathcal{I}^1_{s}, \mathcal{I}^2_{s}$). We denote the dominant predictor as $\mathcal{F}_t(.)$ and phantom predictor as $\mathcal{F}_{p}(.)$. \\

\subsubsection{Network Architecture} Figure~\ref{fig:architecture} (left) reveals two key components of the binding network including a \textit{fully convolutional network} encoder and \textit{source separator module} (SSM). Given a mixed image, $\mathcal{I}_{fb}{\in \mathbb{R}^{h\times w \times c}}$, we adopt DeepLabv3~\cite{chen2017rethinking} ($f_{enc}$) to produce a sequence of bottom-up feature maps. The SSM consists of two separate branches: (i) \textit{dominant}, $\mathcal{F}_t(.)$, and (ii) \textit{phantom}, $\mathcal{F}_{p}(.)$. Each branch takes the spatial feature map, $\hat{f}_{b}^i$, produced at the last block, \texttt{res5c}, of $f_{enc}$ as input and produces a dense prediction for the dominant, $\mathcal{S}_{t}$, and the phantom, $\mathcal{S}_p$, image. Next, we append a \textit{feature binding head} (FBH) to generate a final dense prediction of categories for the dominant image. The FBH, $\mathcal{F}_{fb}$, simply concatenates the outputs of source and phantom branches followed by two $1\times 1$ convolution layers with non-linearities (ReLU) to obtain the final dense prediction map, $\mathcal{S}_{fb}$. The intuition behind the FBH is that the phantom branch may produce activations that are correlated with the dominant image, and thus the FBH allows the network to further correct any incorrectly separated features with an additional signal to learn from. Given a mixed image, $\mathcal{I}_{fb}$, the operations can be expressed as:
\begin{gather}
\hat{f}_{b}^i = f_{\textit{enc}}(\mathcal{I}_{fb}), \hspace{0.2cm} \underbrace{\mathcal{S}_t=\mathcal{F}_t(\hat{f}_{b}^i)}_\texttt{dominant}, \hspace{0.2cm}  \underbrace{\mathcal{S}_{p}=\mathcal{F}_{p}(\hat{f}_{b}^i)}_\texttt{phantom}, \\
\underbrace{\mathcal{S}_{fb}=\mathcal{F}_{fb}(\mathcal{S}_t, \mathcal{S}_{p})}_\texttt{binding}.
\end{gather}

\subsubsection{Training the Feature Binding Network}
\blue{The feature binding network produces two dominant predictions, $\mathcal{S}_{fb}$ and $\mathcal{S}_t$, including a phantom prediction, $\mathcal{S}_p$; however, we are principally interested in the final dominant prediction, $\mathcal{S}_{fb}$. More formally, let $\mathcal{I}_{fb}\in{\rm I\!R}^{h\times w \times 3} $ be a training image associated with ground-truth maps ($\mathcal{G}_a$, $\mathcal{G}_b$) in the feature binding setting. To apply supervision on $\mathcal{S}_{fb}$, $\mathcal{S}_t$, and $\mathcal{S}_p$, we upsample them to the size of $\mathcal{G}_a$. Then we define three pixel-wise cross-entropy losses, $\ell_{fb}$, $\ell_{t}$, and $\ell_{p}$, to measure the difference between ($\mathcal{S}_{fb}$, $\mathcal{G}_a$), ($\mathcal{S}_t$, $\mathcal{G}_a$), and ($\mathcal{S}_p$, $\mathcal{G}_b$), respectively. The objective function can be formalized as:}
\begin{gather}
L_{stage1} = \ell_{fb} + \delta \ast \ell_{t} + (1-\delta) \ast \ell_{p} ,
\end{gather}
where $\delta$ is the weight used in to linearly combine images to generate $\mathcal{I}_{fb}$. Note that the network is penalized the most on the final and initial dominant predictions, and places less emphasis on the phantom prediction.

\subsection{Denoising Feature Binding}\label{sec:denoise}
While feature binding and source separation are interesting, the ultimate goal is to see improvement and robustness for standard images. For this reason, we mainly care about improving the overall dense prediction. To accomplish this, we further fine-tune our trained binding model on the standard training set which we call the feature denoising stage. In this stage, as we feed a standard image to the network, the phantom predictor branch, $\mathcal{F}_{ph}$, has no supervisory signal, instead it acts as a regularizer. We propose the following technique to penalize the phantom prediction.\\



\noindent \textbf{Penalize Phantom Activation.} Along with $\ell_{t}$, we propose a loss, $\ell_{\text{PPA}}$, on the phantom prediction to penalize any activation (and suppress phantom signals and interference). The goal here is to push the output of the phantom branch to zero and suppress the phantom. The $\ell_{\text{PPA}}$ loss sums the absolute value of the confidence attached to categories and applies a $\log$ operation to balance the numeric scale with $\ell_{t}$: 
\begin{gather}\label{eq:ppa}
\ell_{\text{PPA}} = \log\sum_{\forall_{i\in h}}\sum_{\forall_{j\in w}}\sum_{\forall_{k\in c}} \sigma (\mathcal{{S}}_{p}), \hspace{0.3cm}
L_{stage2} = \ell_{t} + \ell_{\text{PPA}},
\end{gather}
where $\sigma(\cdot)$ is the ReLU function, which constrains the input to the $\log$ to be a positive value. In \textbf{Stage 1}, $f_{\textit{enc}}$, $\mathcal{F}_t$, $\mathcal{F}_p$, and $\mathcal{F}_{fb}$ are trained in an end-to-end manner. Then, in \textbf{Stage 2}, $f_{\textit{enc}}$, $\mathcal{F}_t$, and $\mathcal{F}_p$ are fine-tuned from the Stage 1 weights.


\section{Experiments}\label{sec:exp}
We first present results on the PASCAL VOC 2012~\cite{everingham2015pascal} semantic segmentation dataset (Sec.~\ref{sec:seg}).
Unless otherwise stated, we use the DeepLabv3~\cite{chen2017rethinking} network without any bells and whistles (e.g., multi-scale processing, conditional random field) as our baseline model. We then show qualitative and quantitative evidence that our proposed mixing techniques improve the network's ability to segment highly occluded objects in complex scenes (Sec.~\ref{sec:occlusion}), as well as objects found in out-of-context scenarios (Sec.~\ref{sec:context}). Throughout the experiments, we compare our methods to recent mixing strategies, CutMix~\cite{yun2019cutmix} and Mixup~\cite{zhang2017mixup}. Mixup and CutMix did not explicitly design their strategies for dense labeling; however, in CutMix, the authors use CutMix and MixUp for image localization and object detection tasks, so we view their strategies as a general data augmentation technique. Next, we evaluate the robustness of our methods to a variety of adversarial attacks (Sec.~\ref{sec:adver}). \blue{We further apply our co-occurrence based image blending strategy for salient object detection tasks and compare the results with existing techniques (Sec.~\ref{sec:SOD})}. Finally, we conduct an extensive ablation study (Sec.~\ref{sec:ablation}) to better tease out the underlying mechanisms giving performance boosts by evaluating the various image blending strategies and network architectures. 

\subsection{Implementation Details} We implement our proposed feature binding networks using PyTorch~\cite{paszke2017automatic}. 
We apply bilinear interpolation to upsample the predicted segmentation map before the losses are calculated. The \textit{feature binding} networks are trained using stochastic gradient descent for 50 epochs with momentum of 0.9, weight decay of 0.0005 and the “poly” learning rate policy~\cite{chen2018deeplab} which starts at $2.5e^{-4}$. We use the same strategy during the feature denoising stage of training, but with an initial learning rate of $2.5e^{-5}$. During training, we apply random and center cropping to form 513$\times$513 input images during training and inference, respectively. \blue{For a fair comparison, we implement and train Mixup~\cite{zhang2017mixup} and CutMix~\cite{yun2019cutmix} using the same set of hyper-parameters. We report numbers for the following variants that are described in what follows: \textbf{DeepLabv3 + Feature Binding-CC:} This network applies the categorical clustering based image blending with the DeepLabv3 based feature binding network. \textbf{DeepLabv3 + Feature Binding-CM:} This network uses the co-occurrence matrix based image blending with the DeepLabv3 based feature binding network. \textbf{DeepLabv3 + Mixup:} This network uses the Mixup~\cite{guo2019mixup} technique with the DeepLabv3 network. \textbf{DeepLabv3 + CutMix:} This network applies the CutMix~\cite{yun2019cutmix} technique with the DeepLabv3 network for the task of semantic segmentation.}

\subsection{Dataset and Evaluation Metrics}
\noindent \textbf{PASCAL VOC 2012:} The PASCAL VOC 2012 dataset is considered the most popular semantic segmentation dataset, and includes 20 object categories and a background class. It consists of 1464 training images, 1449 validation images, and 1456 testing images. Following the current common practice~\cite{chen2017rethinking,long15_cvpr,refinenet,chen2018deeplab}, we augment the training set using extra labeled PASCAL VOC images from~\cite{hariharan2011semantic}. We use the standard mean IoU metric to report semantic segmentation performance.

\subsection{Results on Semantic Segmentation}\label{sec:seg}
First, we show the improvements on segmentation accuracy by our methods on the PASCAL VOC 2012 validation dataset. We present a comparison of different baselines and our proposed approaches in Table~\ref{tab:voc2012_val_fcnresnet}.

\blue{As shown in Table~\ref{tab:voc2012_val_fcnresnet}, our image blending based binding approaches improve the overall mIoU more than other approaches~\cite{zhang2017mixup,yun2019cutmix}. Additionally, the co-occurrence based blending technique (DeepLabv3+Binding-CM) marginally outperforms the categorical clustering based strategy (0.8\% vs 1.8\% improvement over the baseline DeepLabv3-ResNet101 method). } \\

We further evaluate our approaches on the PASCAL VOC 2012 test set. Following prior works~\cite{chen2018deeplab,zhao2017pyramid,noh15_iccv}, before evaluating our method on the test set, we first train on the augmented training set followed by fine-tuning on the original trainval set. \blue{As shown in Table~\ref{tab:voc2012_test_fcnresnet}, DeepLabv3 with categorical clustering based feature binding network achieves 80.5\% mIoU which outperforms the baseline. Additionally, co-occurrence based feature binding network achieves 81.1\% mIoU which marginally outperforms the baselines and the categorical clustering based feature binding network.}

\begin{table}[t]
\begin{center}
		\setlength\tabcolsep{1.2pt}
		\def\arraystretch{1.25}
		\resizebox{0.48\textwidth}{!}{
		    \begin{tabular}{c|l|cc}
				\specialrule{1.2pt}{1pt}{1pt}
				Backbone&\multicolumn{1}{c|}{Method}&  mIoU (\%) \\
				\specialrule{1.2pt}{1pt}{1pt}
				\multirow{5}{*}{Res50}&DeepLabv3-ResNet50~\cite{chen2017rethinking} & 75.1\\
				&DeepLabv3 + Mixup~\cite{zhang2017mixup} & 73.6 \\
				& DeepLabv3 + CutMix~\cite{yun2019cutmix} & 75.1 \\
				&\textbf{DeepLabv3 + Binding-CC} &  75.7 \\
					&\textbf{DeepLabv3 + Binding-CM} & \textbf{76.2}\\

				\specialrule{1.2pt}{1pt}{1pt}
				\multirow{5}{*}{Res101}&DeepLabv3-ResNet101~\cite{chen2017rethinking} & 77.1\\
				&DeepLabv3 + Mixup~\cite{zhang2017mixup} & 76.2 \\
				&DeepLabv3 + CutMix~\cite{yun2019cutmix} & 76.7 \\
				&\textbf{DeepLabv3 + Binding-CC} &  77.9\\
					&\textbf{DeepLabv3 + Binding-CM} & \textbf{78.9}\\
				\specialrule{1.2pt}{1pt}{1pt} 

			\end{tabular}
		}

		\caption{\blue{Quantitative comparisons on PASCAL VOC 2012 val set. Our co-occurrence based feature binding network outperforms the other mixing based techniques.}}
	    \label{tab:voc2012_val_fcnresnet}
	    	\end{center}
\end{table}

\begin{table}[t]
\begin{center}
		\setlength\tabcolsep{1.2pt}
		\def\arraystretch{1.25}
		\resizebox{0.48\textwidth}{!}{
		    \begin{tabular}{l|cc}
				\specialrule{1.2pt}{1pt}{1pt}
				\multicolumn{1}{c|}{Method}&  mIoU (\%) \\
				\specialrule{1.2pt}{1pt}{1pt}
			
				DeepLabv3-ResNet101~\cite{chen2017rethinking} & 79.3\\
				DeepLabv3 + Mixup~\cite{zhang2017mixup} & 78.9 \\
				DeepLabv3 + CutMix~\cite{yun2019cutmix} &  80.2\\
				\textbf{DeepLabv3 + Feature Binding (CC)} & 80.5 \\
				\textbf{DeepLabv3 + Feature Binding (CM)} & \textbf{81.1}\\
				\specialrule{1.2pt}{1pt}{1pt} 

			\end{tabular}
		}

		\caption{Quantitative comparisons of various mixing techniques on PASCAL VOC 2012 test set.}
	    \label{tab:voc2012_test_fcnresnet}
	    	\end{center}
\end{table}

\begin{figure*}
	\begin{center}

		\resizebox{0.99\textwidth}{!}{
		\def\arraystretch{0.3}
        \setlength\tabcolsep{0.5pt}
		    
		\begin{tabular}{*{7}{c}}		
				
				\includegraphics[width=0.15\textwidth]{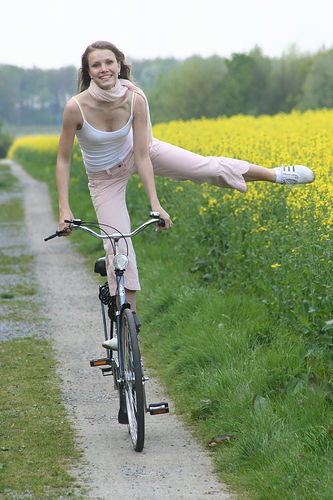}&
				\includegraphics[width=0.15\textwidth]{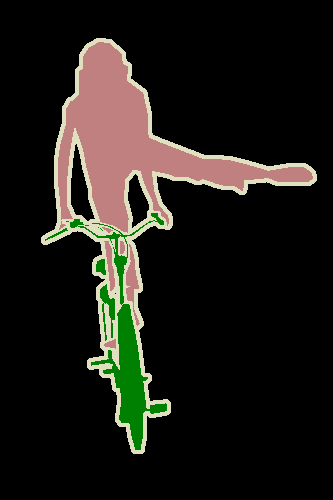}&
				\includegraphics[width=0.15\textwidth]{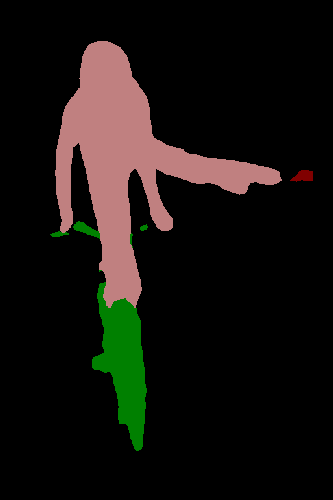}&
				\includegraphics[width=0.15\textwidth]{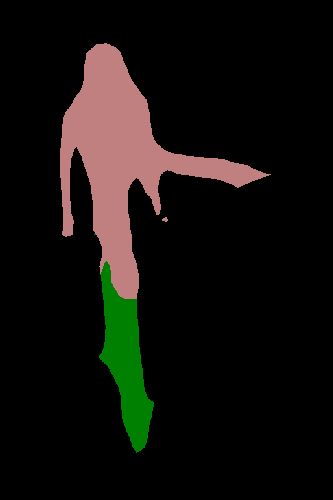}&
				\includegraphics[width=0.15\textwidth]{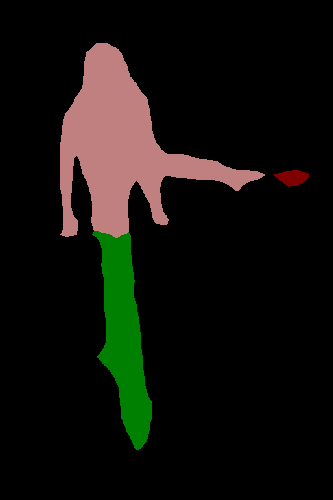}&
				\includegraphics[width=0.15\textwidth]{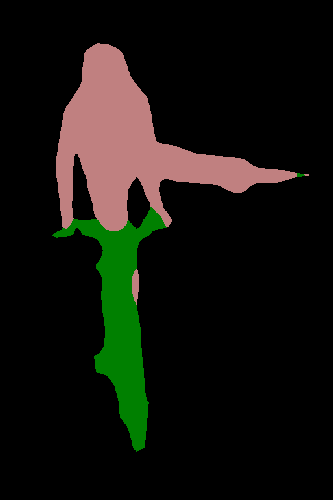}&
				\includegraphics[width=0.15\textwidth]{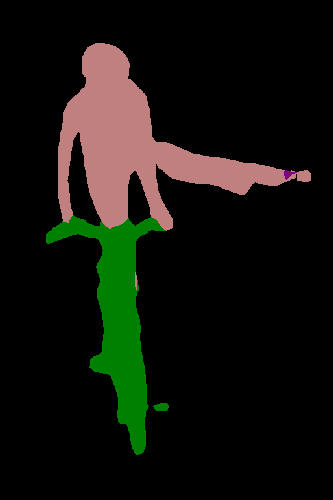}\\
				
				\includegraphics[width=0.15\textwidth]{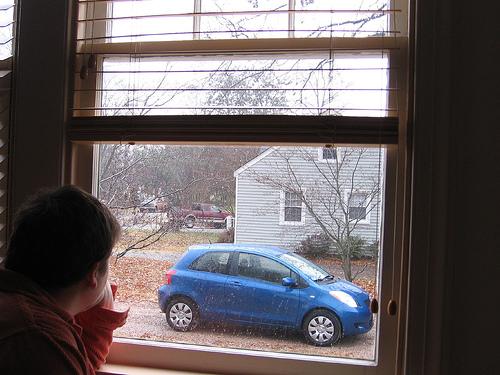}&
				\includegraphics[width=0.15\textwidth]{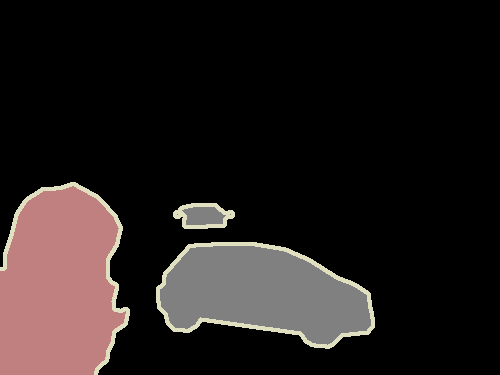}&
				\includegraphics[width=0.15\textwidth]{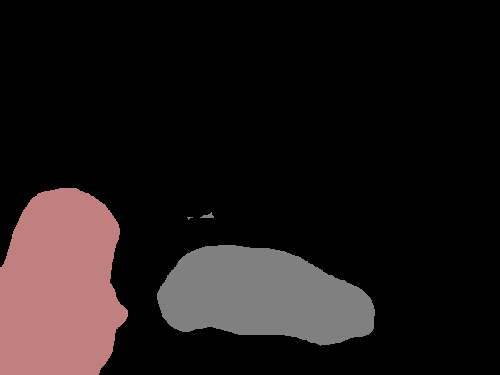}&
				\includegraphics[width=0.15\textwidth]{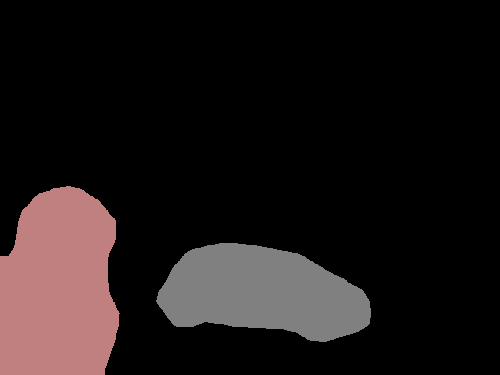}&
				\includegraphics[width=0.15\textwidth]{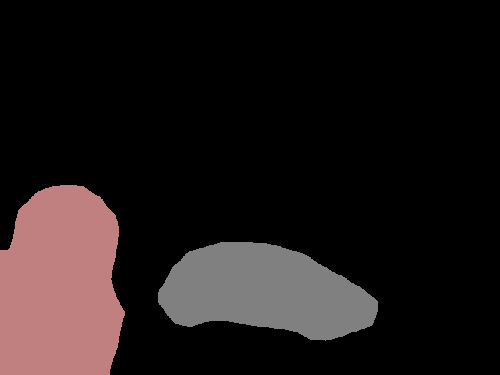}&
				\includegraphics[width=0.15\textwidth]{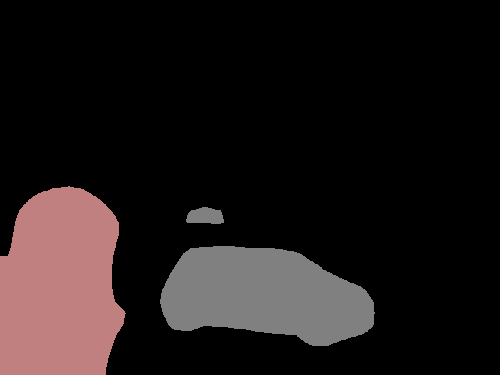}&
				\includegraphics[width=0.15\textwidth]{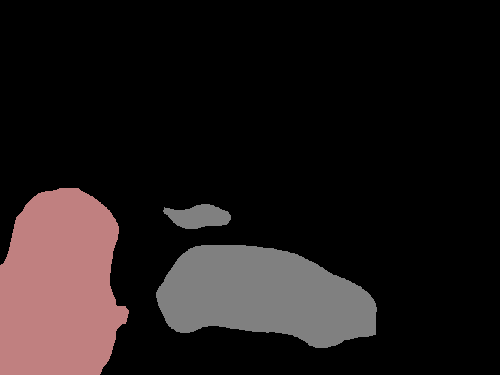}\\
				
				\includegraphics[width=0.15\textwidth]{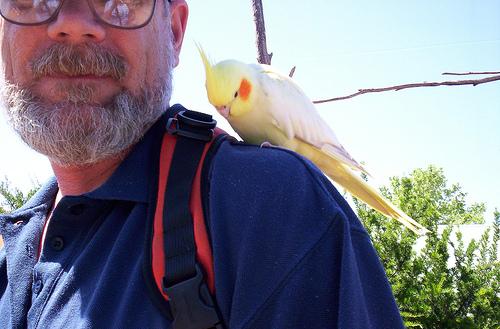}&
				\includegraphics[width=0.15\textwidth]{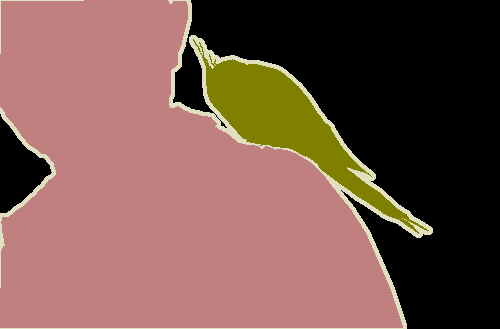}&
				\includegraphics[width=0.15\textwidth]{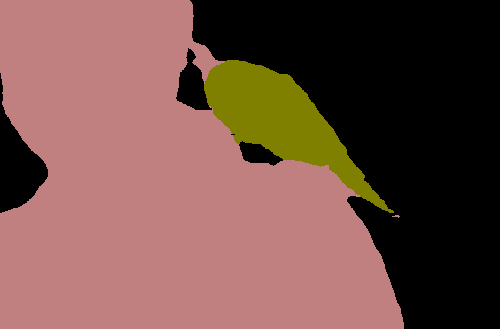}&
				\includegraphics[width=0.15\textwidth]{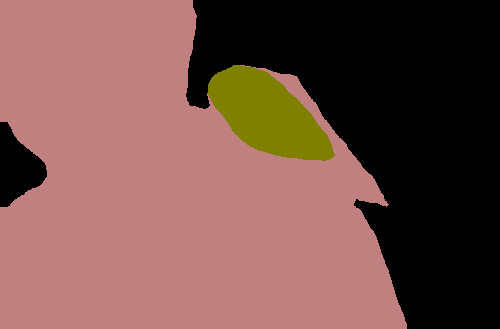}&
				\includegraphics[width=0.15\textwidth]{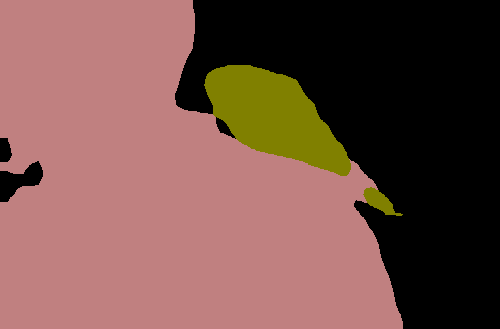}&
				\includegraphics[width=0.15\textwidth]{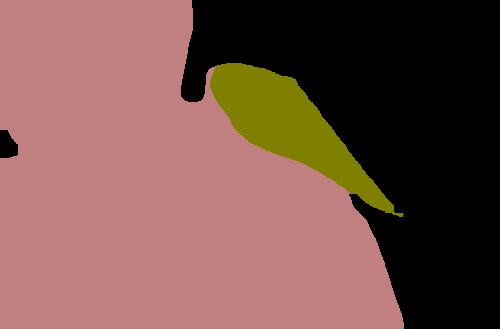}&
				\includegraphics[width=0.15\textwidth]{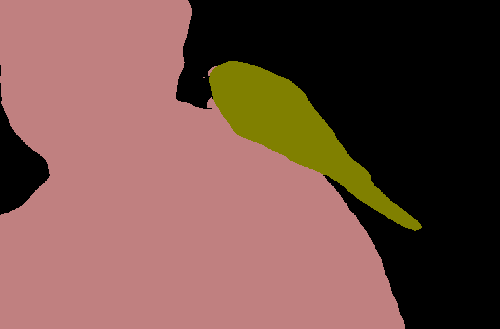}\\

				\includegraphics[width=0.15\textwidth]{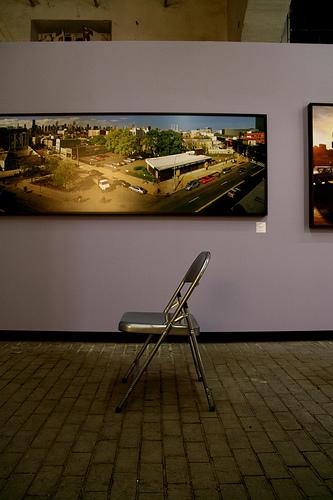}&
				\includegraphics[width=0.15\textwidth]{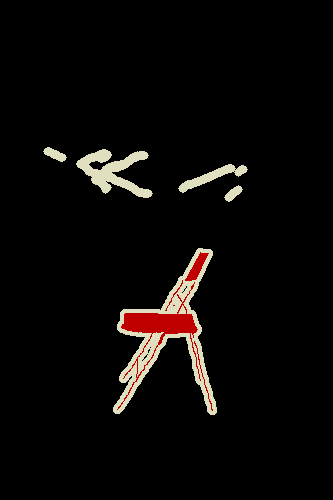}&
				\includegraphics[width=0.15\textwidth]{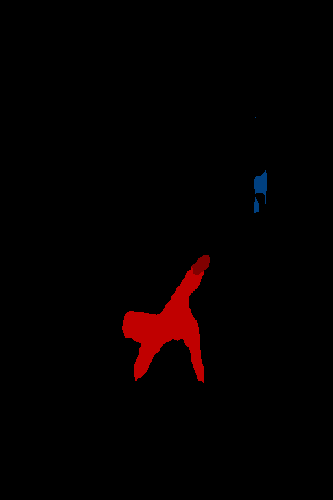}&
				\includegraphics[width=0.15\textwidth]{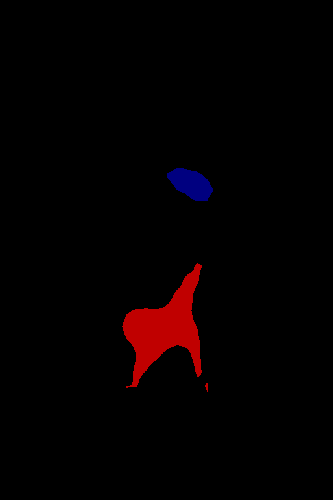}&
				\includegraphics[width=0.15\textwidth]{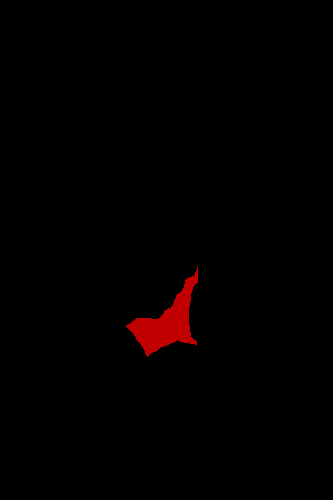}&
				\includegraphics[width=0.15\textwidth]{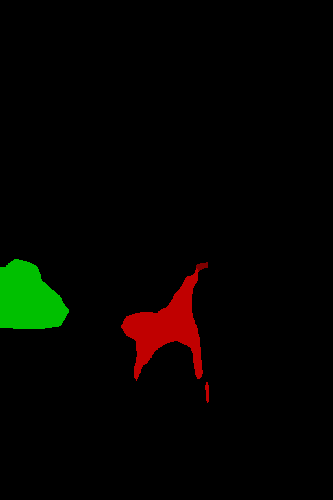}&
				\includegraphics[width=0.15\textwidth]{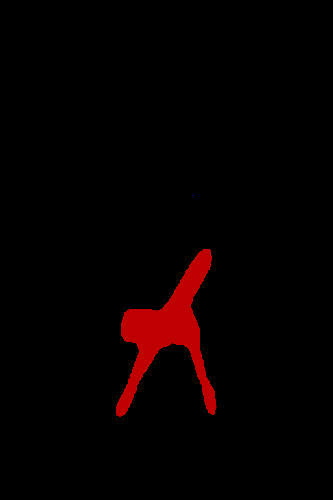}\\
				
				\includegraphics[width=0.15\textwidth]{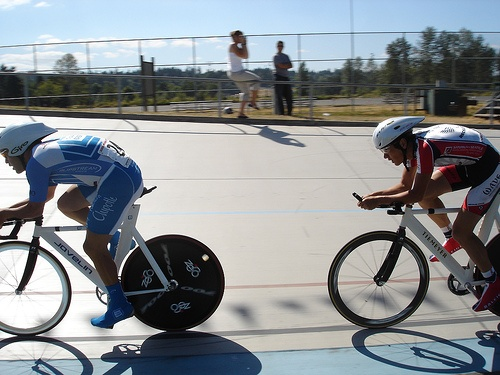}&
				\includegraphics[width=0.15\textwidth]{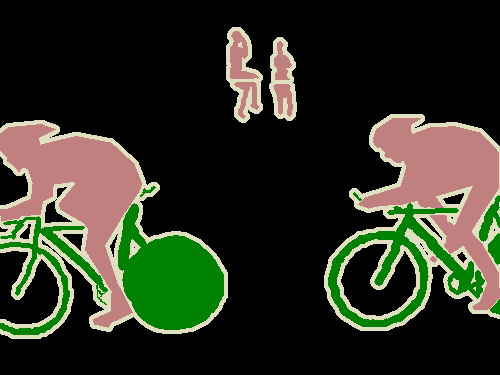}&
				\includegraphics[width=0.15\textwidth]{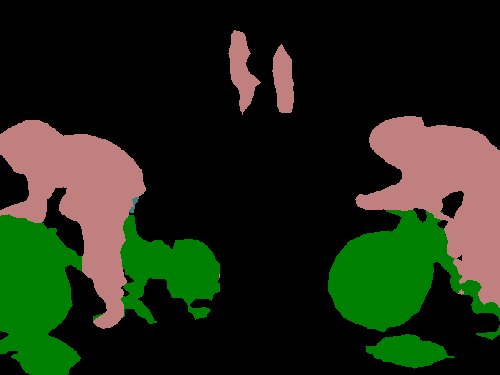}&
				\includegraphics[width=0.15\textwidth]{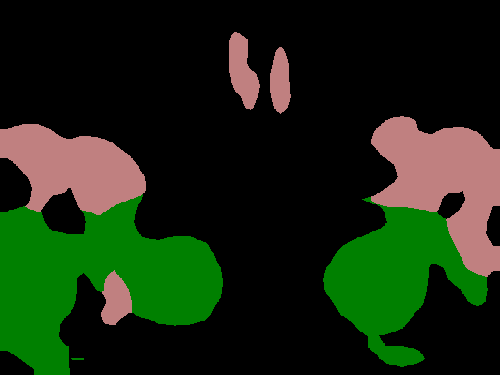}&
				\includegraphics[width=0.15\textwidth]{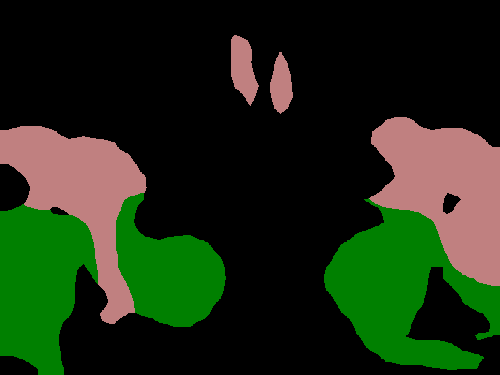}&
					\includegraphics[width=0.15\textwidth]{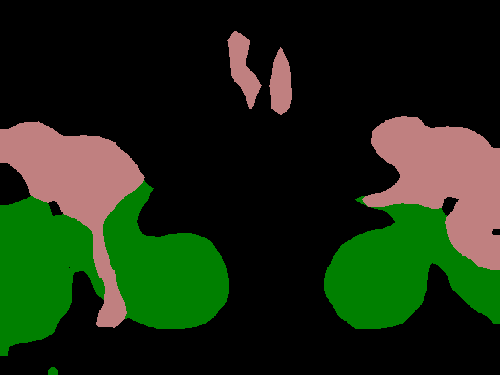}&
				\includegraphics[width=0.15\textwidth]{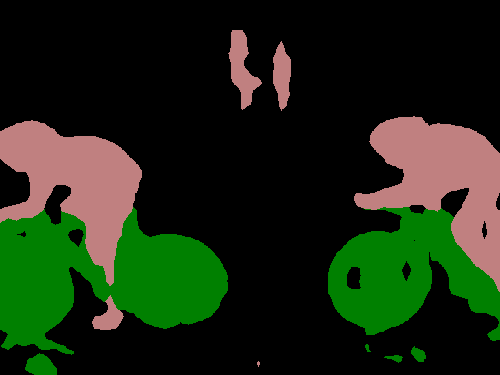}\\
				
				\includegraphics[width=0.15\textwidth]{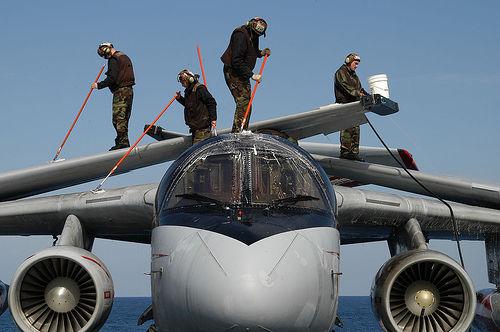}&
				\includegraphics[width=0.15\textwidth]{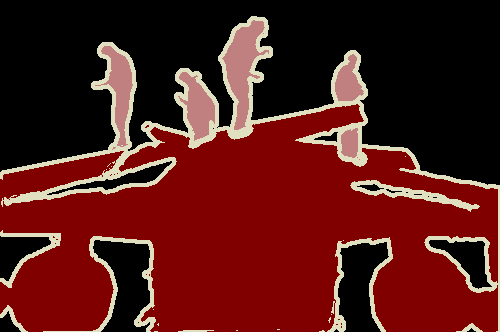}&
				\includegraphics[width=0.15\textwidth]{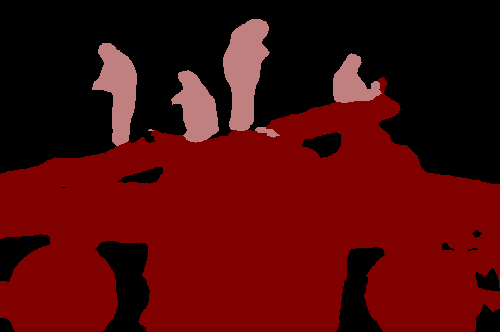}&
				\includegraphics[width=0.15\textwidth]{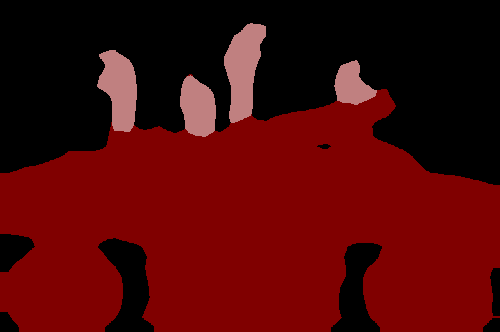}&
				\includegraphics[width=0.15\textwidth]{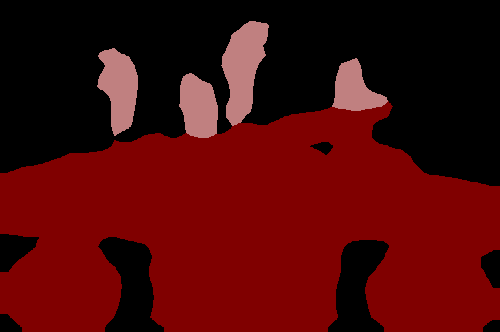}&
				
				\includegraphics[width=0.15\textwidth]{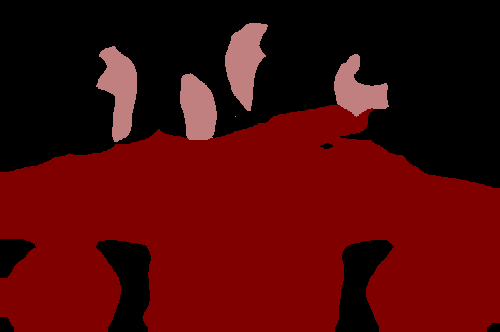}&
				\includegraphics[width=0.15\textwidth]{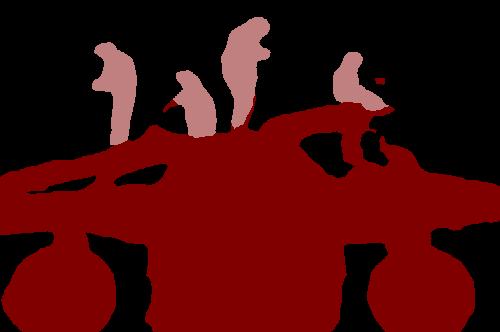}\\

				\includegraphics[width=0.15\textwidth]{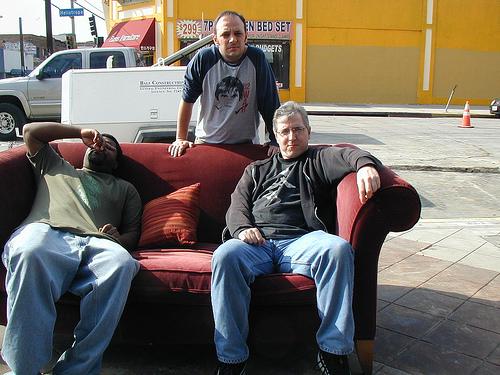}&
				\includegraphics[width=0.15\textwidth]{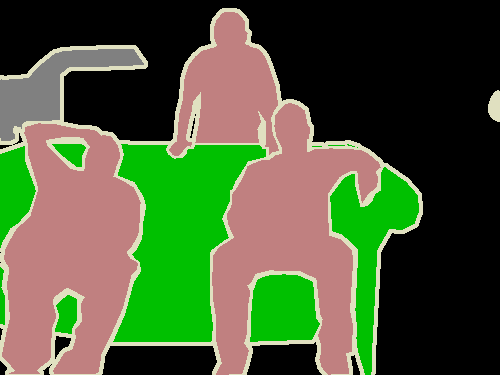}&
				\includegraphics[width=0.15\textwidth]{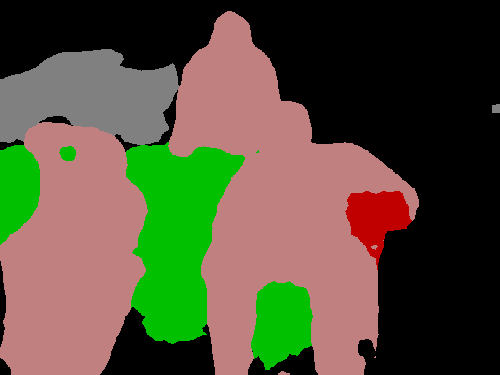}&
				\includegraphics[width=0.15\textwidth]{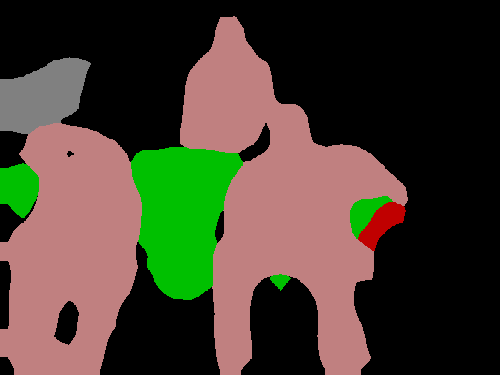}&
				\includegraphics[width=0.15\textwidth]{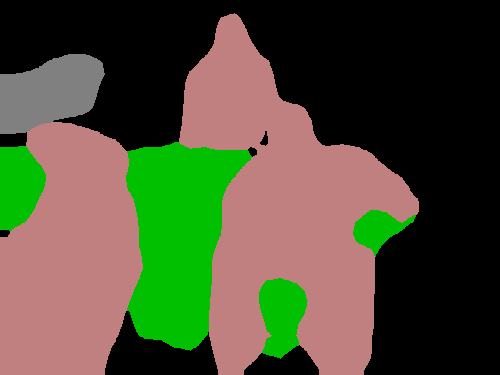}&
				\includegraphics[width=0.15\textwidth]{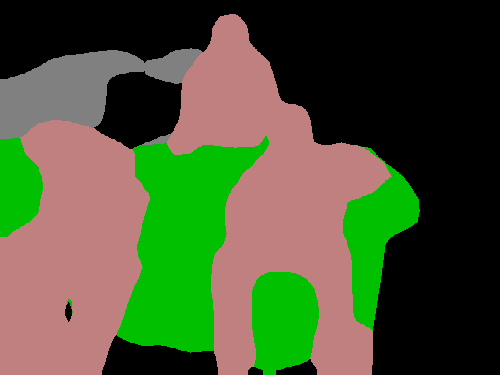}&
				\includegraphics[width=0.15\textwidth]{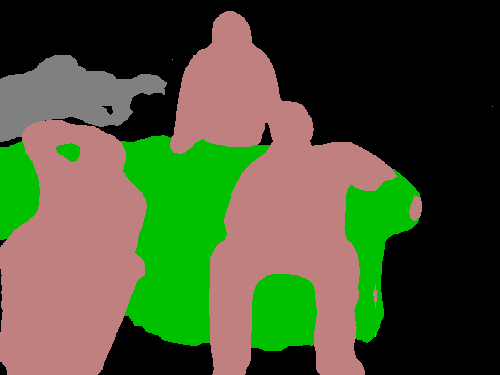}\\
				
				\includegraphics[width=0.15\textwidth]{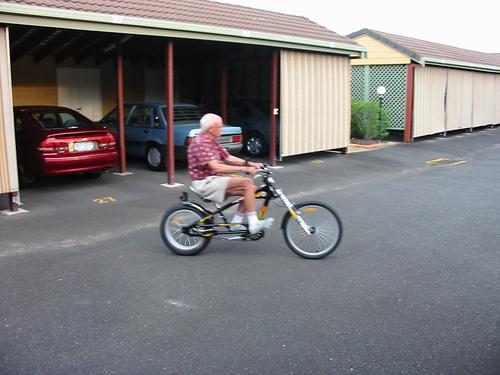}&
				\includegraphics[width=0.15\textwidth]{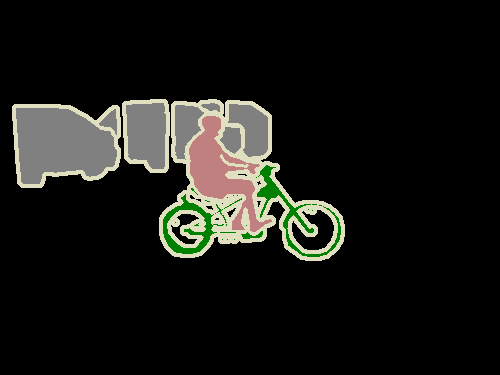}&
				\includegraphics[width=0.15\textwidth]{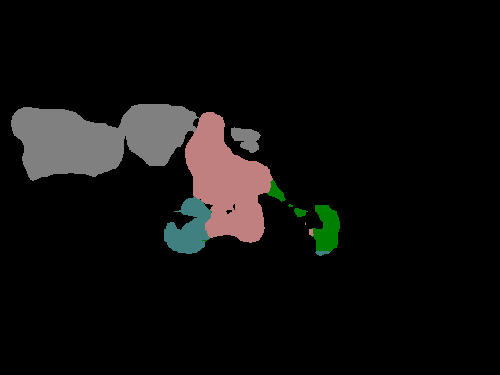}&
				\includegraphics[width=0.15\textwidth]{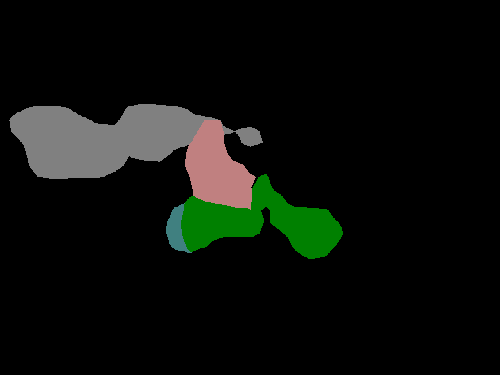}&
				\includegraphics[width=0.15\textwidth]{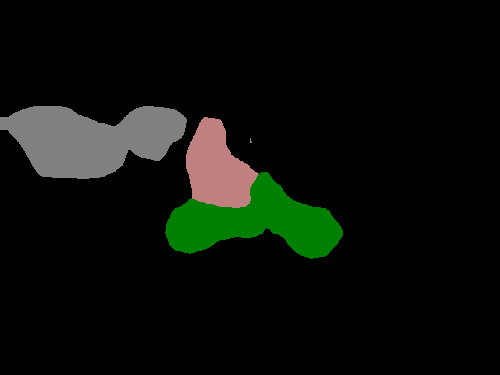}&
				
					\includegraphics[width=0.15\textwidth]{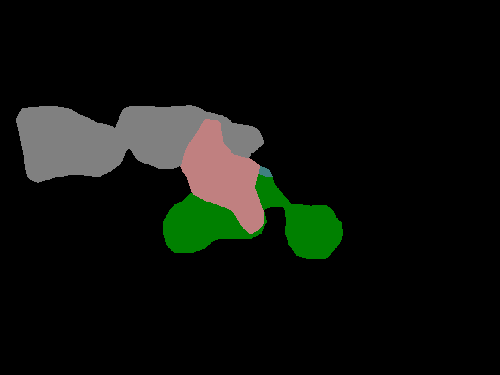}&
				\includegraphics[width=0.15\textwidth]{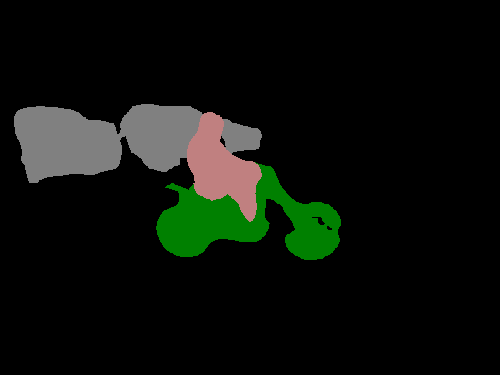}\\
				
				\includegraphics[width=0.15\textwidth]{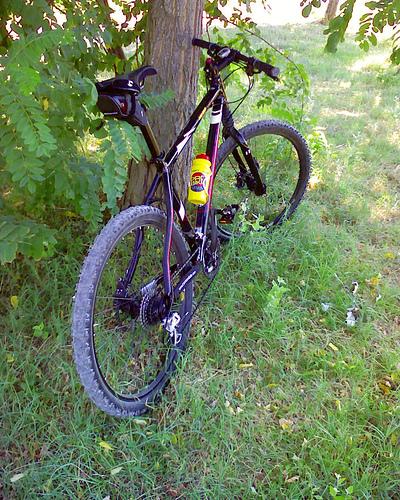}&
				\includegraphics[width=0.15\textwidth]{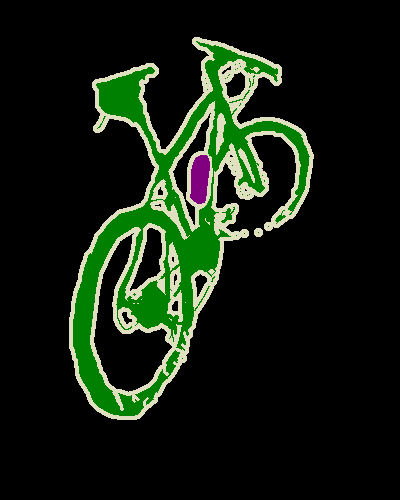}&
				\includegraphics[width=0.15\textwidth]{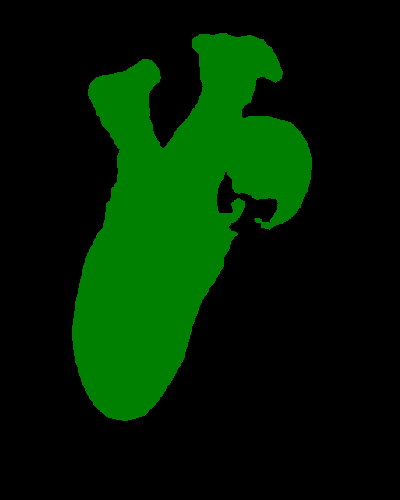}&
				\includegraphics[width=0.15\textwidth]{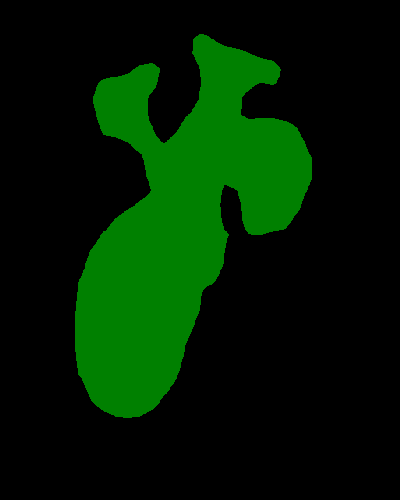}&
				\includegraphics[width=0.15\textwidth]{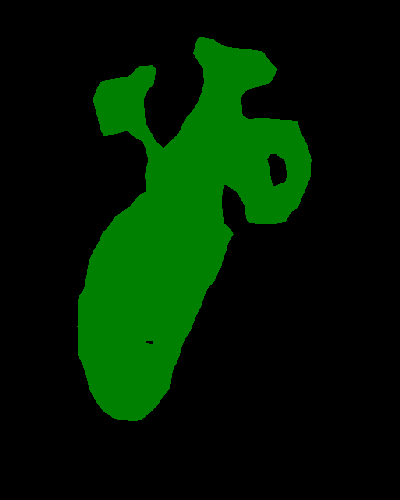}&
				
					\includegraphics[width=0.15\textwidth]{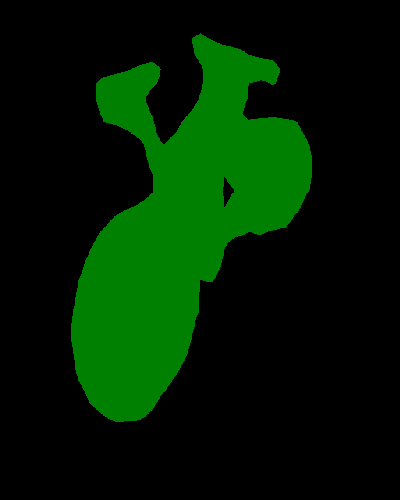}&
				\includegraphics[width=0.15\textwidth]{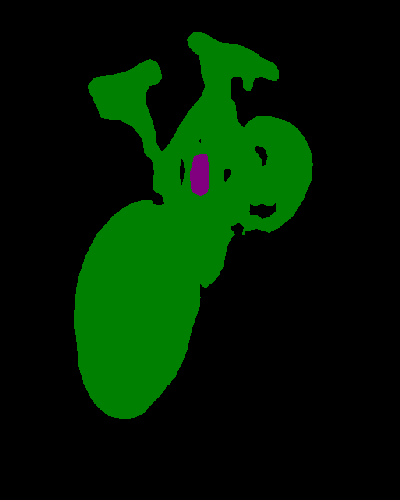}\\

				\includegraphics[width=0.15\textwidth]{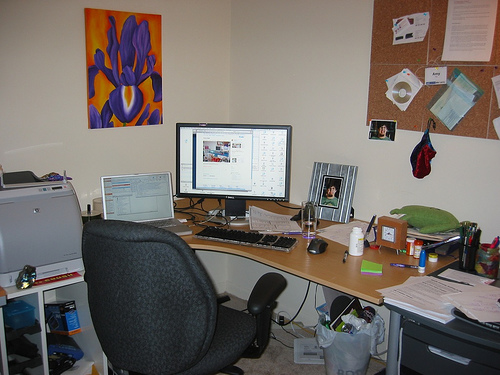}&
				\includegraphics[width=0.15\textwidth]{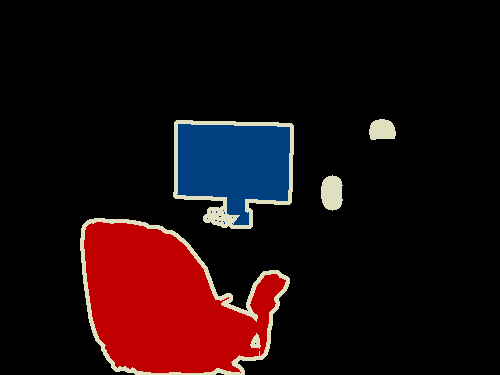}&
				\includegraphics[width=0.15\textwidth]{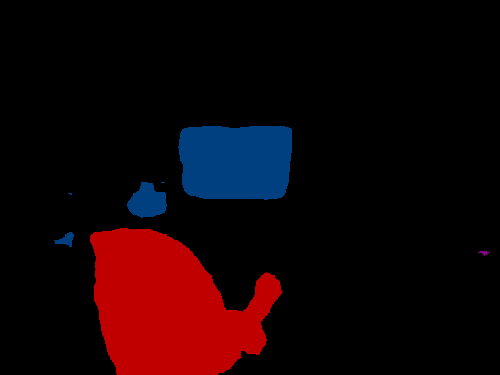}&
				\includegraphics[width=0.15\textwidth]{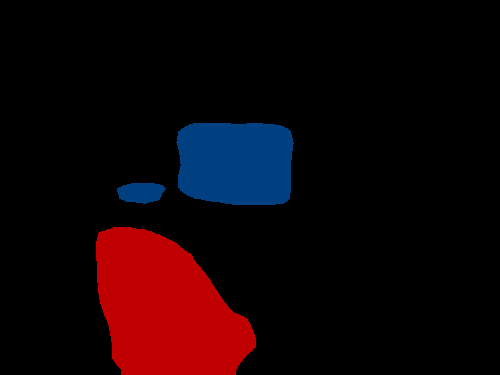}&
				\includegraphics[width=0.15\textwidth]{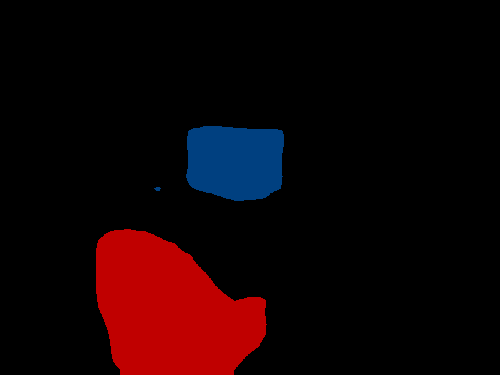}&
				
					\includegraphics[width=0.15\textwidth]{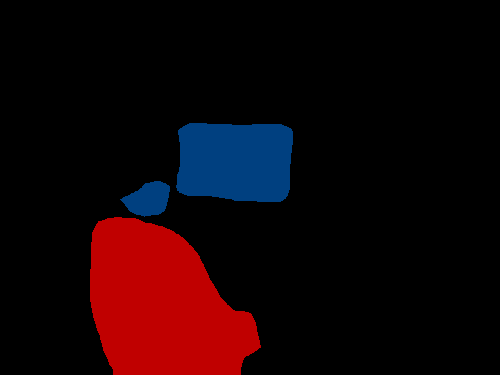}&
				\includegraphics[width=0.15\textwidth]{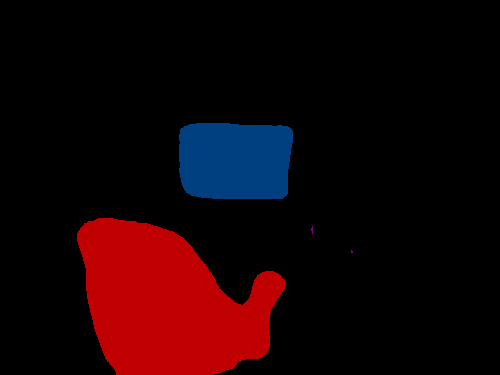}\\				
				Image & GT& DeepLabv3~\cite{chen2017rethinking} &CutMix~\cite{yun2019cutmix}& Mixup~\cite{guo2019mixup} & Binding-CC (ours)&Binding-CM (ours) \\
			\end{tabular}}
			\caption{Qualitative results on the PASCAL VOC 2012 val set.}
			\label{fig:val_pascal}
				\end{center}
\end{figure*}

Sample predictions of our methods and the baselines are shown in Fig.~\ref{fig:val_pascal}. As shown in Fig.~\ref{fig:val_pascal}, our proposed blending based feature binding networks are very effective in capturing more distinct features for labeling occluded objects and plays a critical role in separating different semantic objects more accurately. Note the ability of our methods to segment scenes with a high degree of occlusion (see second last row in Fig.~\ref{fig:val_pascal}), thin overlapping regions (see top row), or complex interaction between object categories (see 6$^{th}$ row). While other methods identify the dominant categories correctly, they often fail to relate the activations of smaller occluding features to the correct categorical assignments.
\begin{table*}
	\centering
	\def\arraystretch{1.25}
		\setlength\tabcolsep{3.5pt}

	\resizebox{0.99\textwidth}{!}{
		\begin{tabular}{l|cc cc ccc  c ccc}
			\specialrule{1.2pt}{1pt}{1pt}\	
			&  \multicolumn{2}{c}{\textit{Occlusion}} && \multicolumn{4}{c}{\textit{Number of Objects}}&& \multicolumn{3}{c}{\textit{Number of \textit{Unique} Objects}} \\
			\cline{2-3} \cline{5-8} \cline{10-12} 
			& \textbf{1-Occ} & \textbf{All-Occ} &&  \textbf{1-Obj} & \textbf{2-Obj
			}& \textbf{3-Obj} & \textbf{4-Obj} &&    \textbf{2-Obj}& \textbf{3-Obj} & \textbf{4-Obj}  \\
			\hline
			\hspace{0.8cm} \# of Images & 1128	& 538 && 695 & 318&	167&	99 &&		375&	121&	23 \\
			\specialrule{1.2pt}{1pt}{1pt}
			
			DeepLabv3 &75.5&	74.9&&	74.6&	74.8&	76.0&	70.0&&		72.5&	\text{63.5}&	62.1 \\
            
           DeepLabv3 + Mixup & 75.4&	72.3&&	77.9&	74.3&	71.7&	68.3&&		72.0	&58.1&	59.2\\
           
           DeepLabv3 + CutMix & 76.4&	74.3&&	78.3&	75.4&	73.0&	70.0&&		72.3&	60.1&	59.6\\
		   
		   \textbf{DeepLabv3 + Binding (CC)} & \text{77.9}&	\textbf{76.1}&&	\textbf{80.7}&	\textbf{77.2}&	\text{75.6}&	\text{70.0}&&		\text{74.0}&	61.5&	\text{62.0}\\
		   
		    
		    \textbf{DeepLabv3 + Binding (CM)} & \textbf{78.7} & \textbf{76.1} && \textbf{80.7} & \textbf{77.2} &\textbf{77.9} & \textbf{72.1} && \textbf{75.4}	&\textbf{65.6}&	\textbf{63.6}\\
			\specialrule{1.2pt}{1pt}{1pt}
		\end{tabular}}
		\caption{Results on complex scenes in terms of mIoU, evaluated using various subsets from the PASCAL VOC 2012 val set. \textit{Occlusion:} Number of occluded objects in the image. \textit{Number of Objects:} Number of objects in the image. \textit{Number of Unique Objects:} Unique object classes contained in the image.}
		\label{table:occlusion}
\end{table*}


\begin{figure*} [t]
	  \begin{center}
		\setlength\tabcolsep{1.1pt}
		\def\arraystretch{0.5}
		\resizebox{0.86\textwidth}{!}{
			\begin{tabular}{*{6}{c}}

				\includegraphics[width=0.15\textwidth]{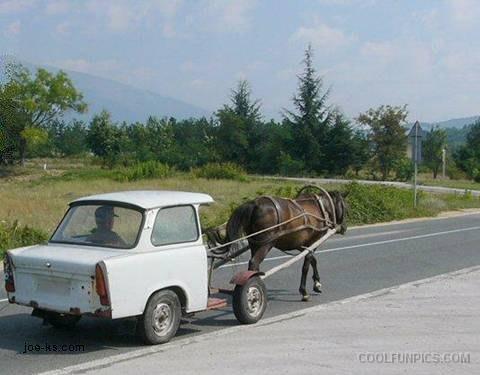}&
				\includegraphics[width=0.15\textwidth]{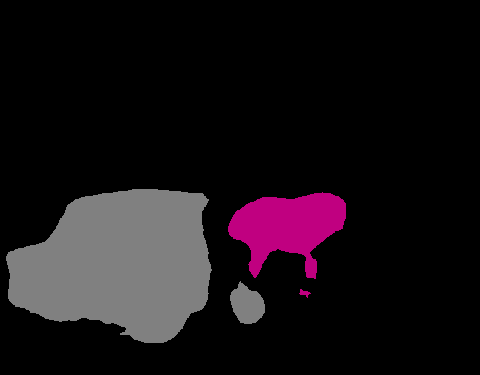}&
				\includegraphics[width=0.15\textwidth]{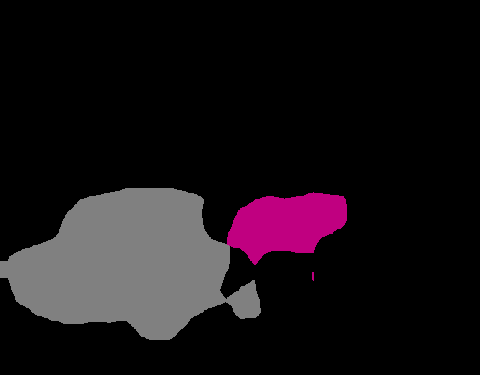}&
				\includegraphics[width=0.15\textwidth]{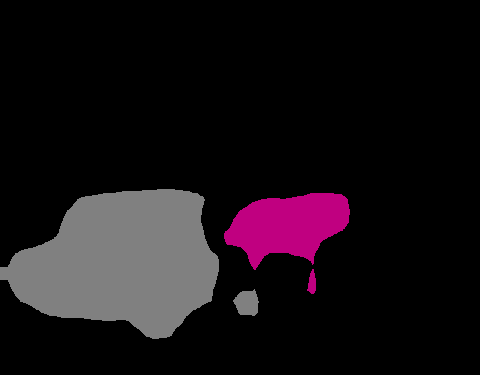}&
				\includegraphics[width=0.15\textwidth]{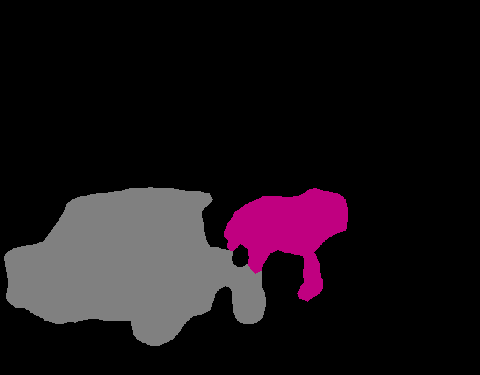}&
				\includegraphics[width=0.15\textwidth]{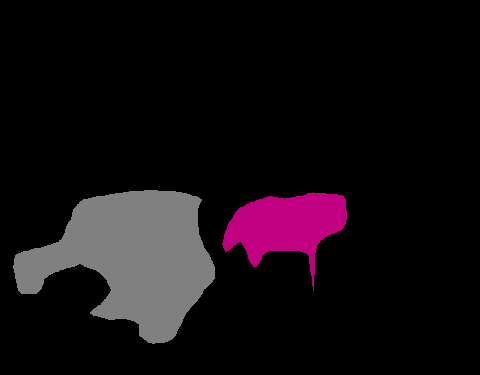}\\

				\includegraphics[width=0.15\textwidth]{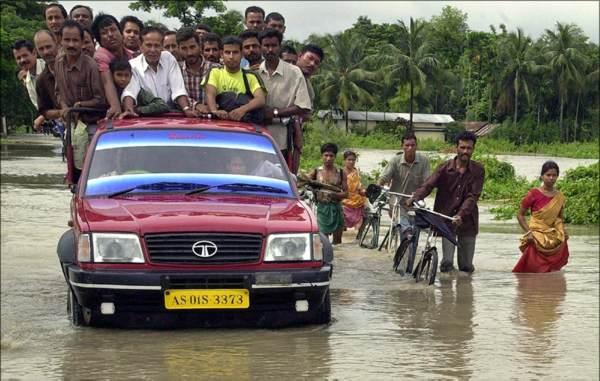}&
				\includegraphics[width=0.15\textwidth]{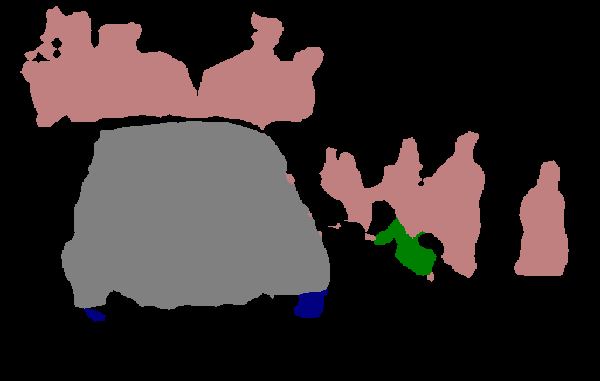}&
				\includegraphics[width=0.15\textwidth]{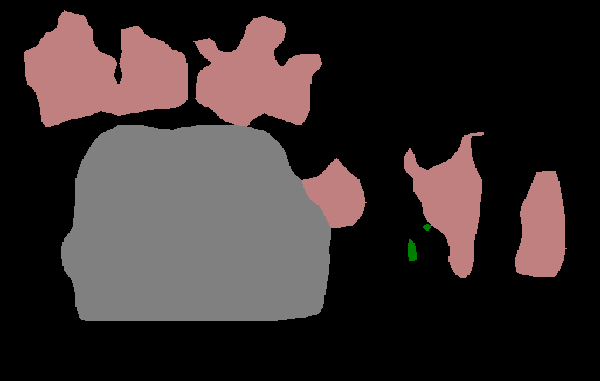}&
				\includegraphics[width=0.15\textwidth]{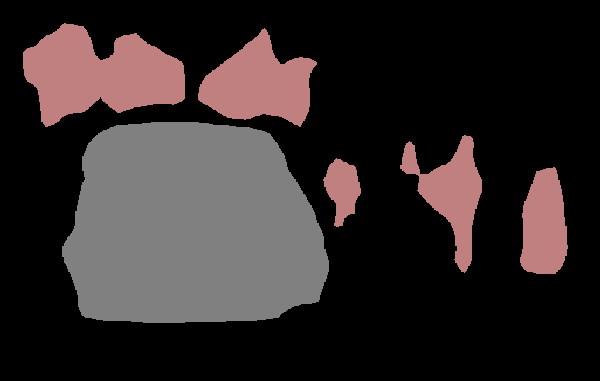}&
					\includegraphics[width=0.15\textwidth]{images/out_of_context/cutmix/im117.png}&
				\includegraphics[width=0.15\textwidth]{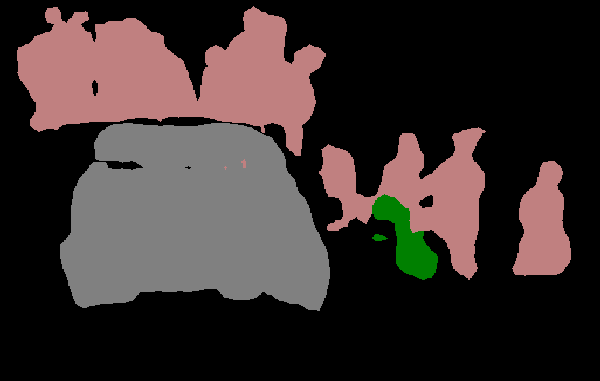}\\
				
				\includegraphics[width=0.15\textwidth]{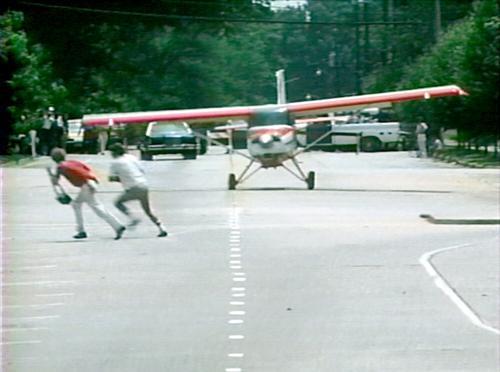}&
				\includegraphics[width=0.15\textwidth]{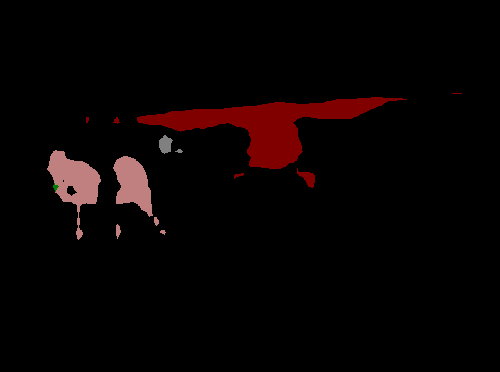}&
				\includegraphics[width=0.15\textwidth]{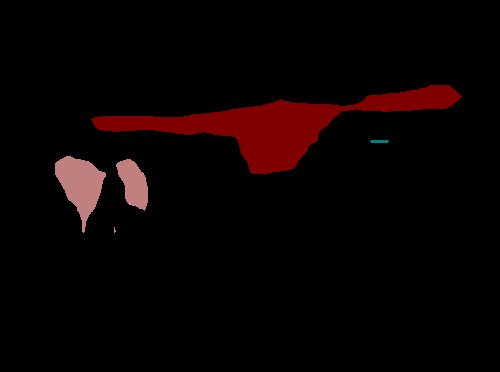}&
				\includegraphics[width=0.15\textwidth]{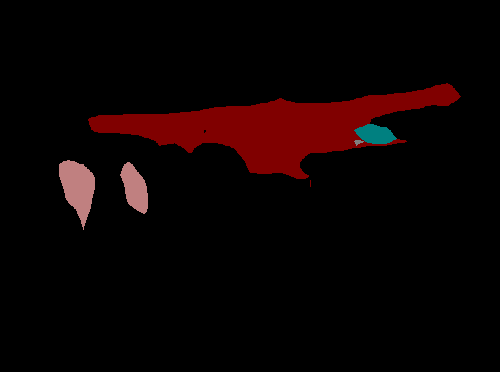}&
					\includegraphics[width=0.15\textwidth]{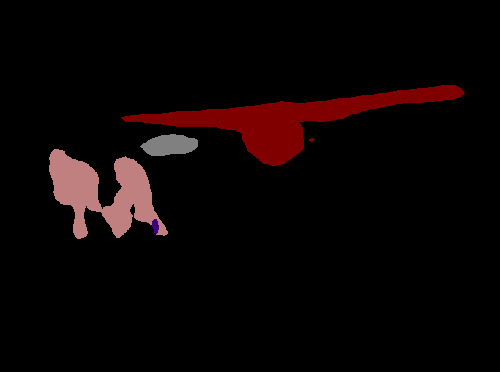}&
				\includegraphics[width=0.15\textwidth]{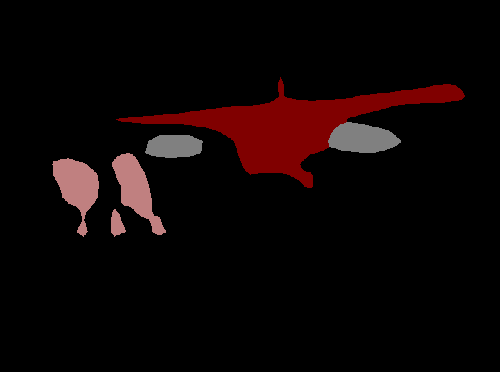}\\
				
				\includegraphics[width=0.15\textwidth]{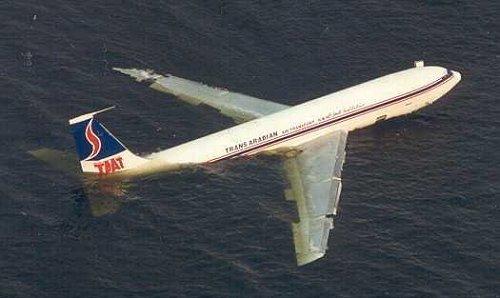}&
				\includegraphics[width=0.15\textwidth]{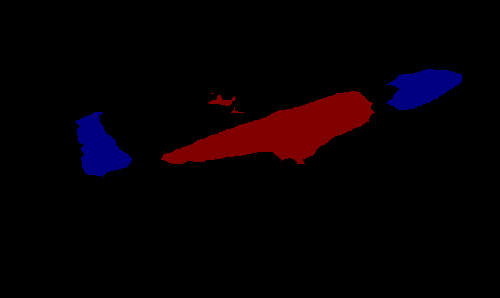}&
				\includegraphics[width=0.15\textwidth]{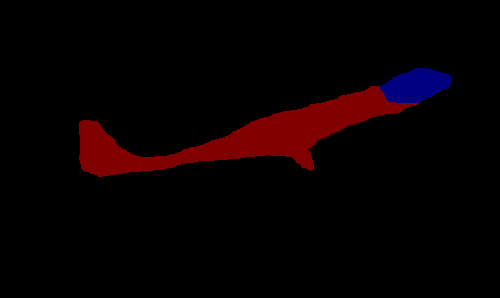}&
				\includegraphics[width=0.15\textwidth]{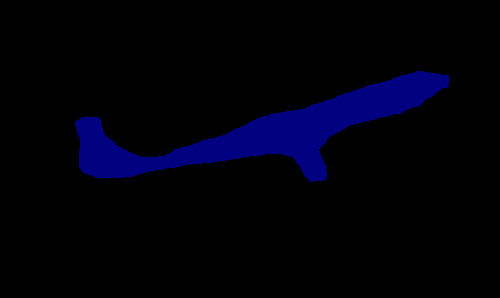}&
				\includegraphics[width=0.15\textwidth]{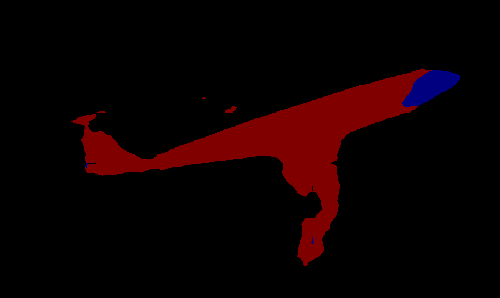}&
				\includegraphics[width=0.15\textwidth]{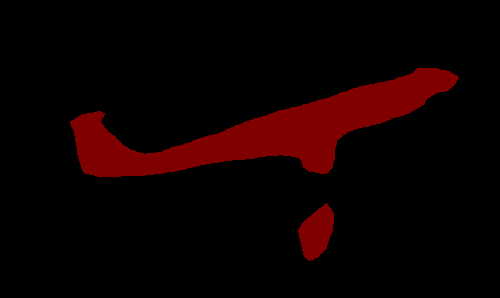}\\
				
				\includegraphics[width=0.15\textwidth]{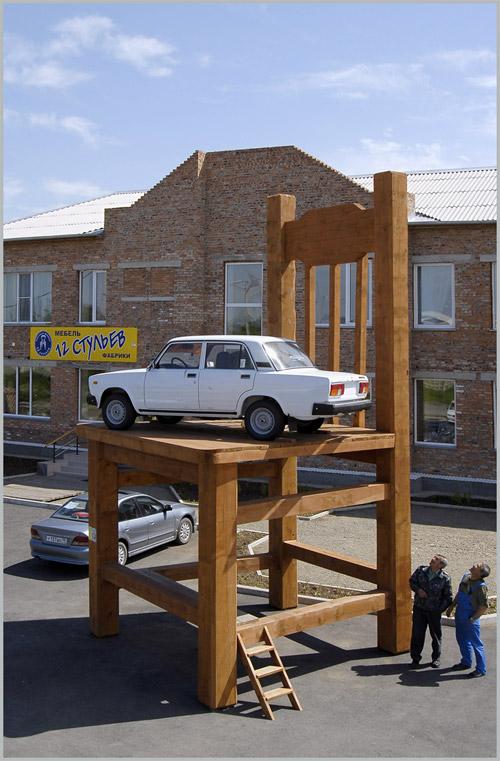}&
				\includegraphics[width=0.15\textwidth]{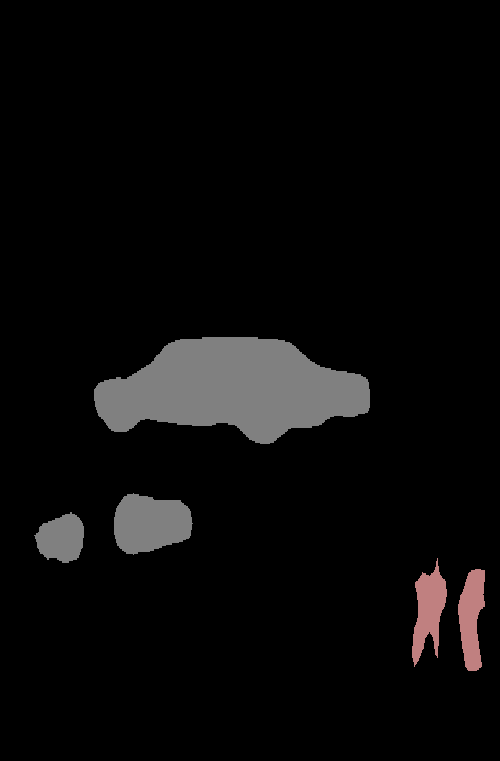}&
				\includegraphics[width=0.15\textwidth]{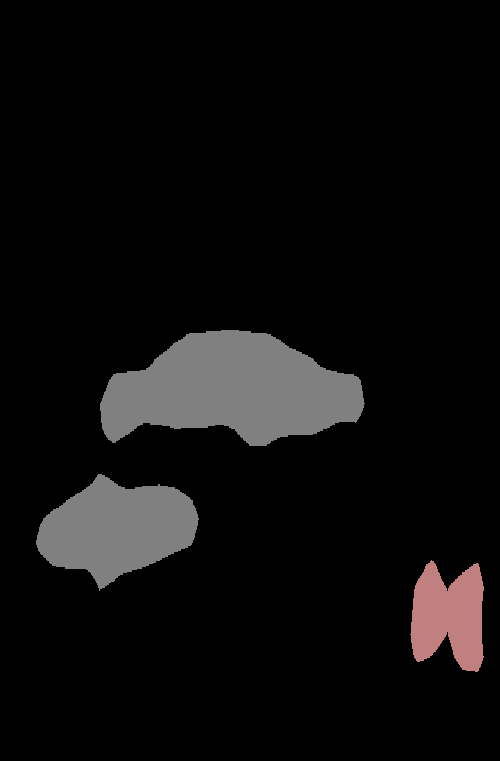}&
				\includegraphics[width=0.15\textwidth]{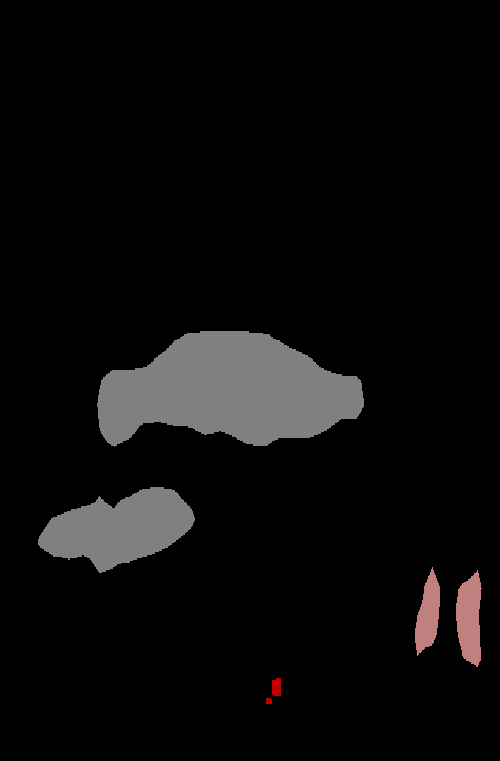}&
				\includegraphics[width=0.15\textwidth]{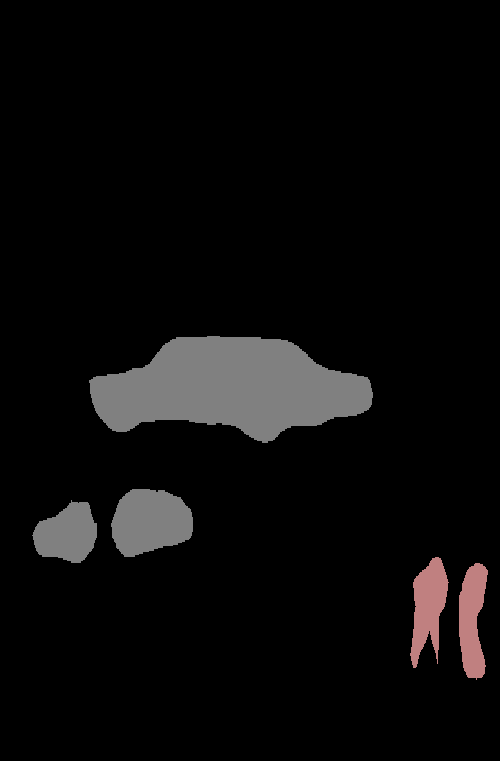}&
				\includegraphics[width=0.15\textwidth]{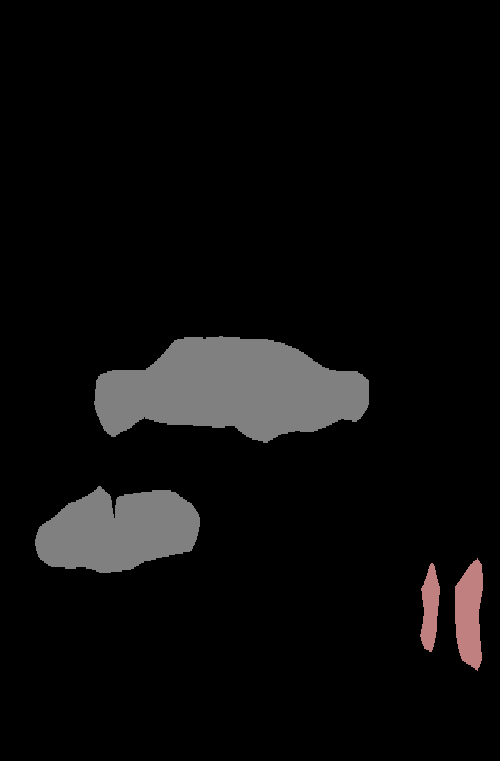}\\
				
					\includegraphics[width=0.15\textwidth, height=1.2cm]{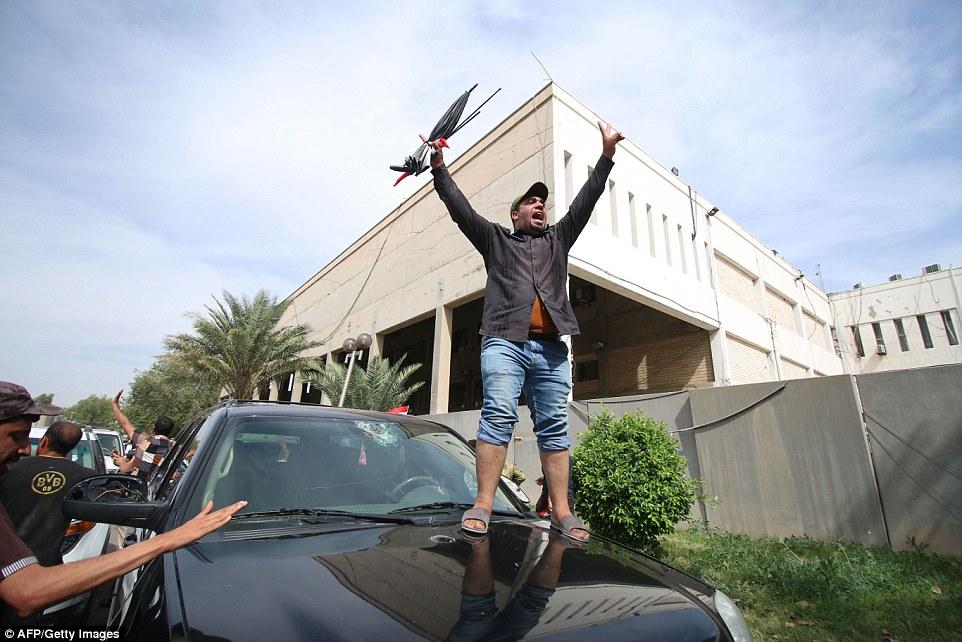}&
				\includegraphics[width=0.15\textwidth, height=1.2cm]{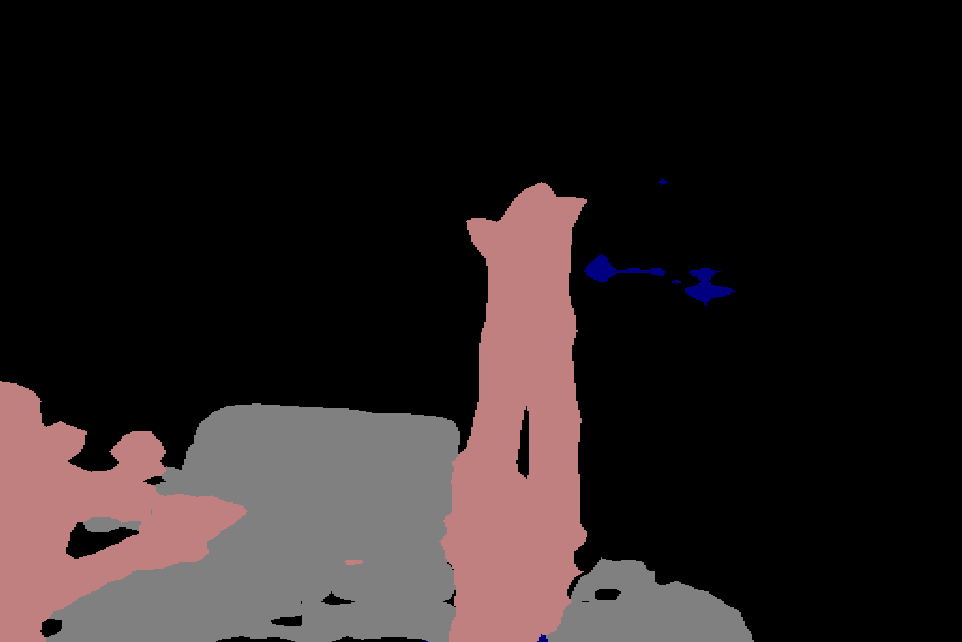}&
				\includegraphics[width=0.15\textwidth, height=1.2cm]{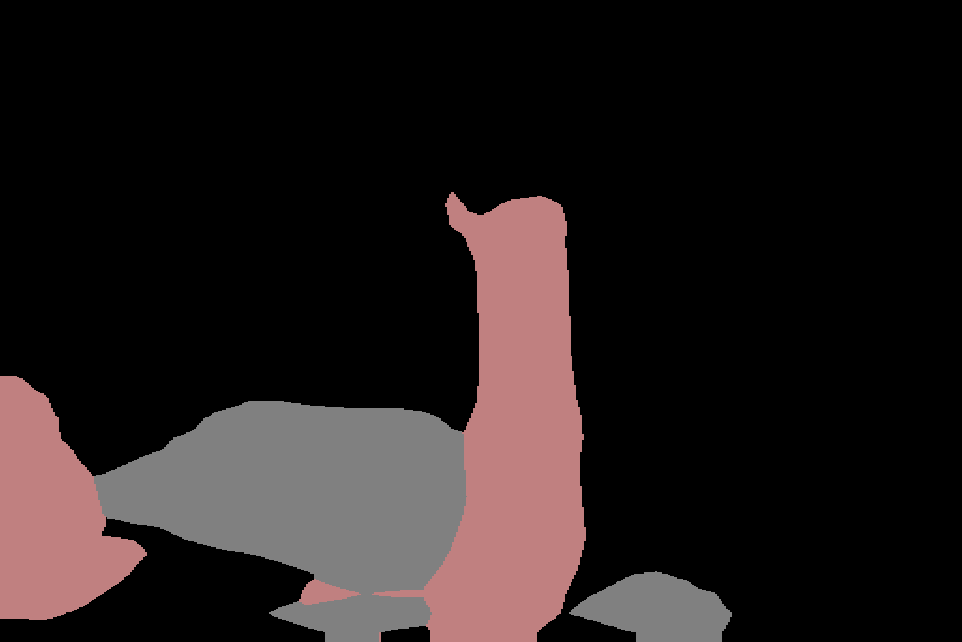}&
				\includegraphics[width=0.15\textwidth, height=1.2cm]{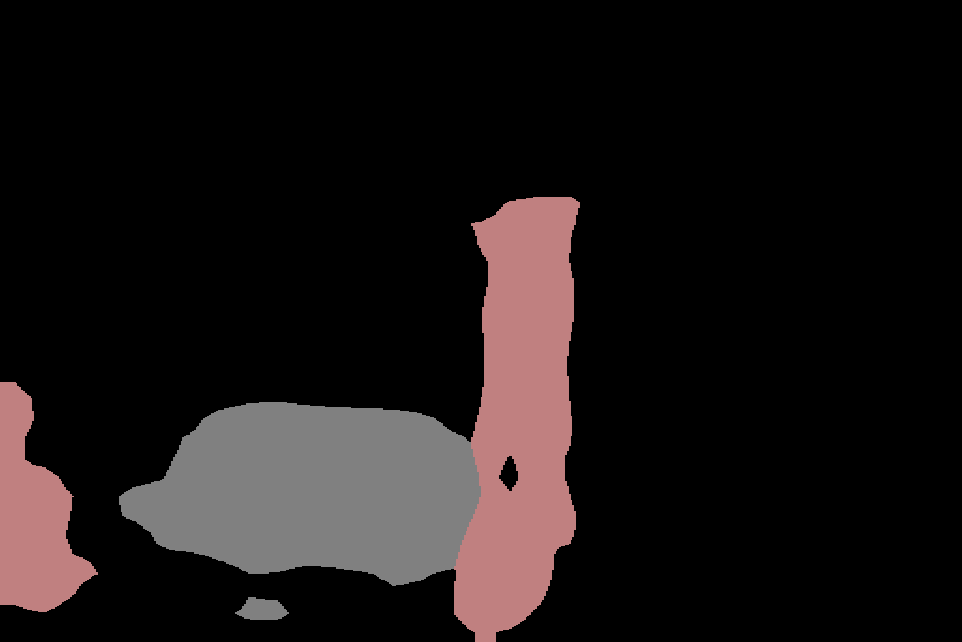}&
				\includegraphics[width=0.15\textwidth, height=1.2cm]{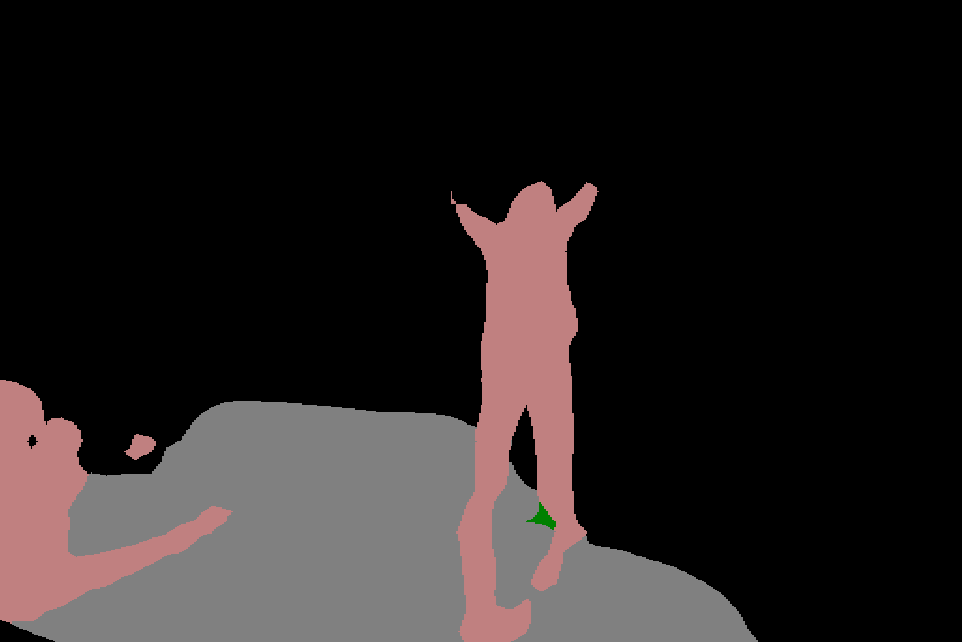}&
				\includegraphics[width=0.15\textwidth, height=1.2cm]{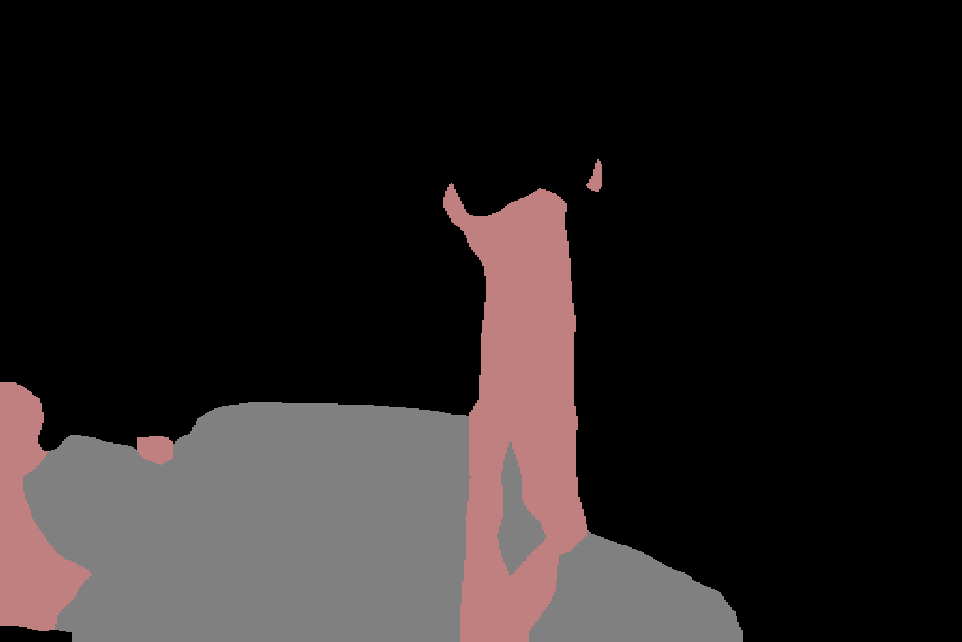}\\
				
					\includegraphics[width=0.15\textwidth]{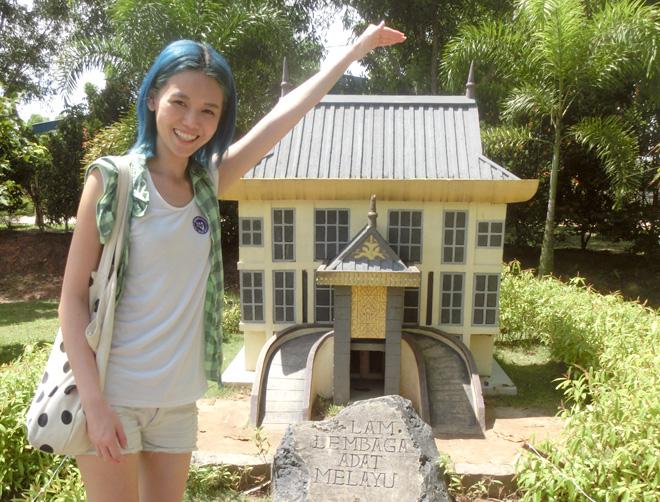}&
				\includegraphics[width=0.15\textwidth]{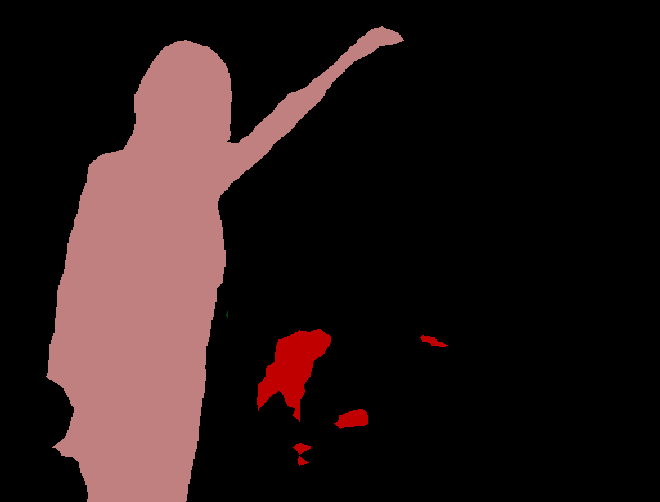}&
				\includegraphics[width=0.15\textwidth]{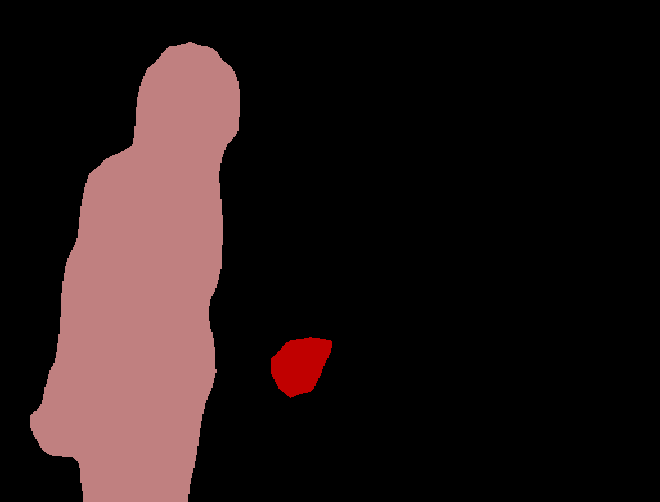}&
				\includegraphics[width=0.15\textwidth]{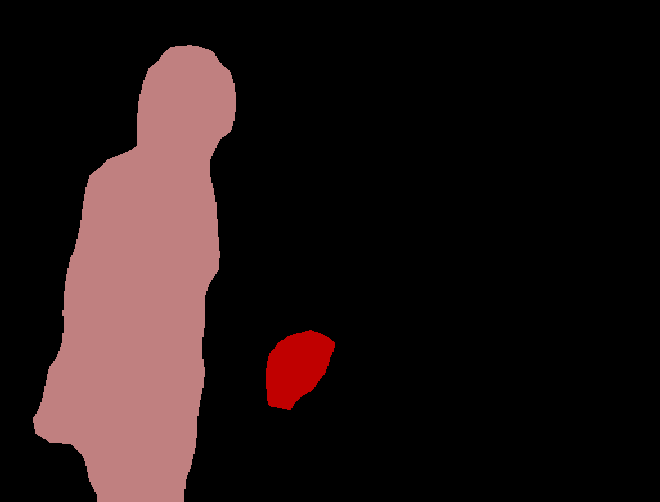}&
					\includegraphics[width=0.15\textwidth]{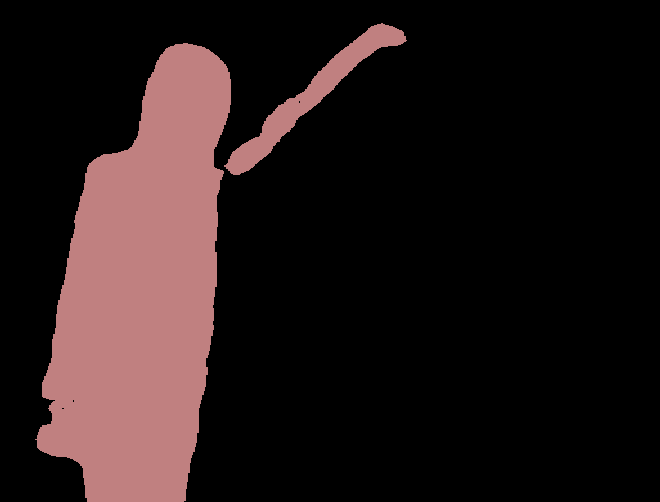}&
				\includegraphics[width=0.15\textwidth]{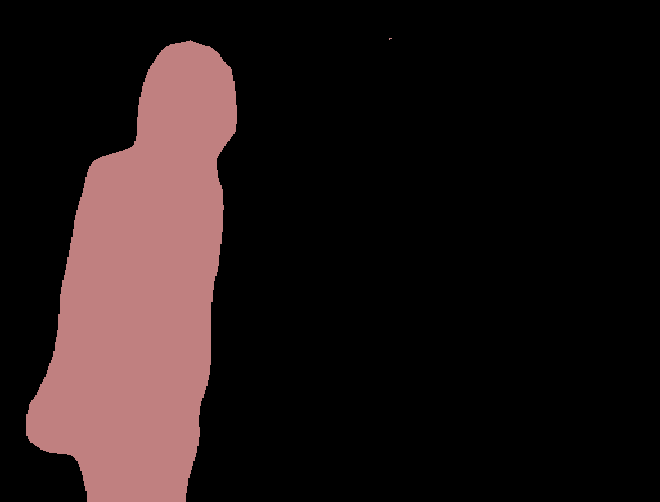}\\
				
					\includegraphics[width=0.15\textwidth]{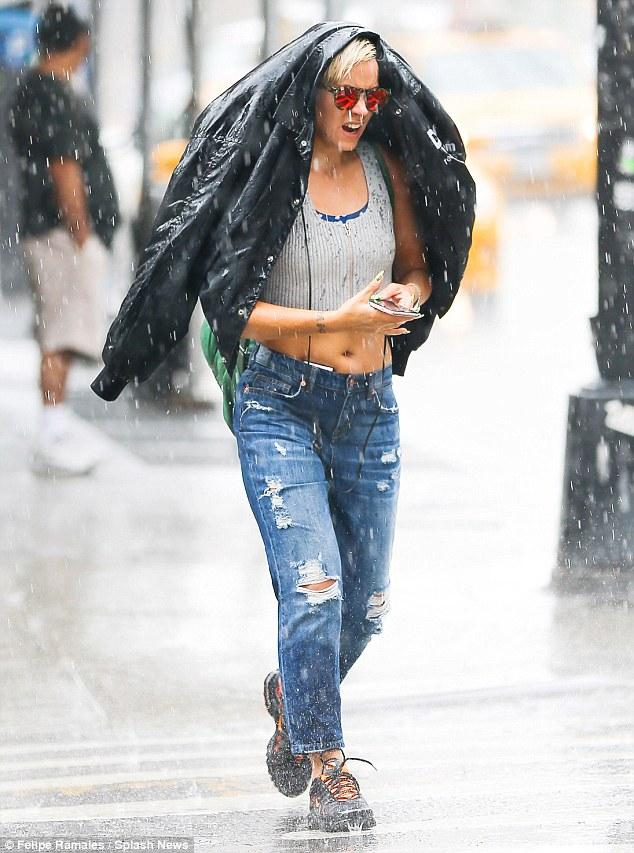}&
				\includegraphics[width=0.15\textwidth]{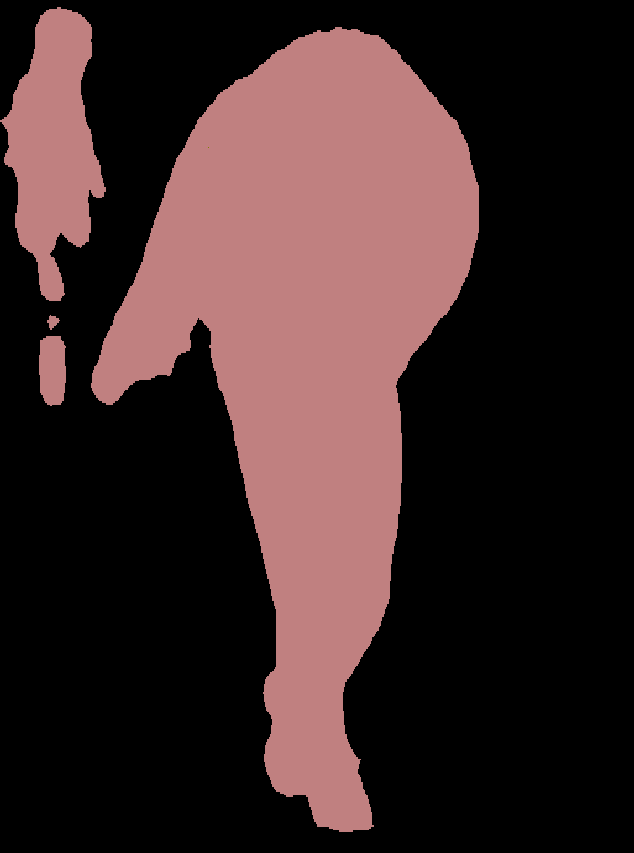}&
				\includegraphics[width=0.15\textwidth]{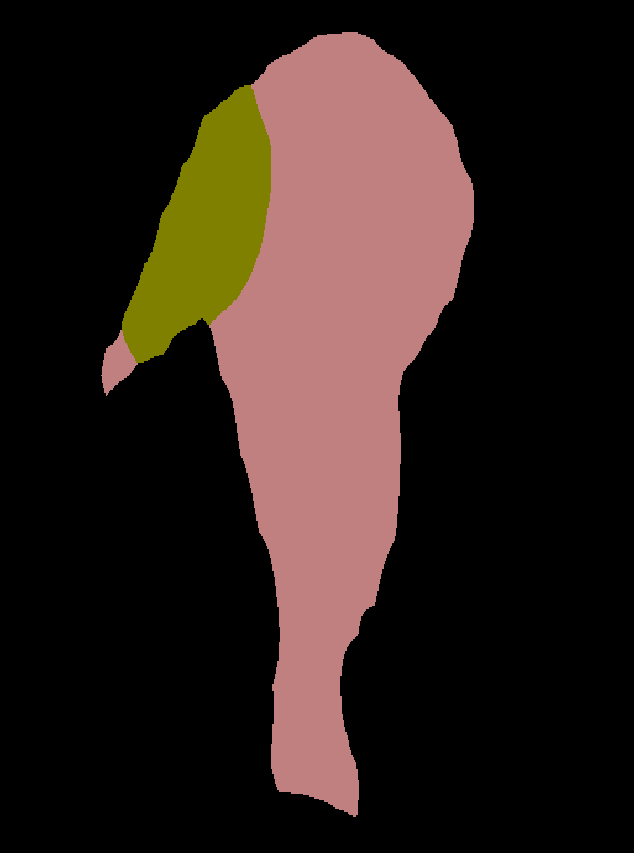}&
				\includegraphics[width=0.15\textwidth]{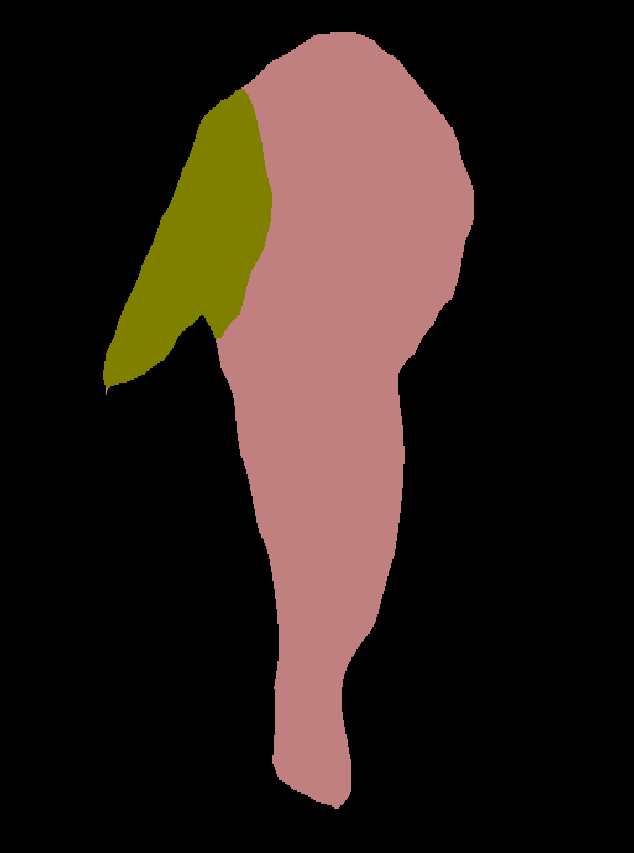}&
					\includegraphics[width=0.15\textwidth]{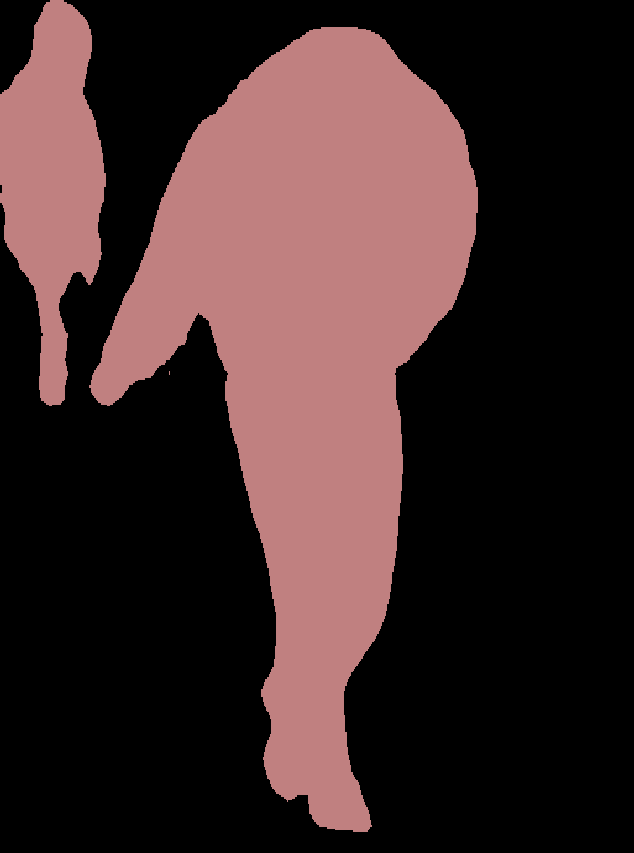}&
				\includegraphics[width=0.15\textwidth]{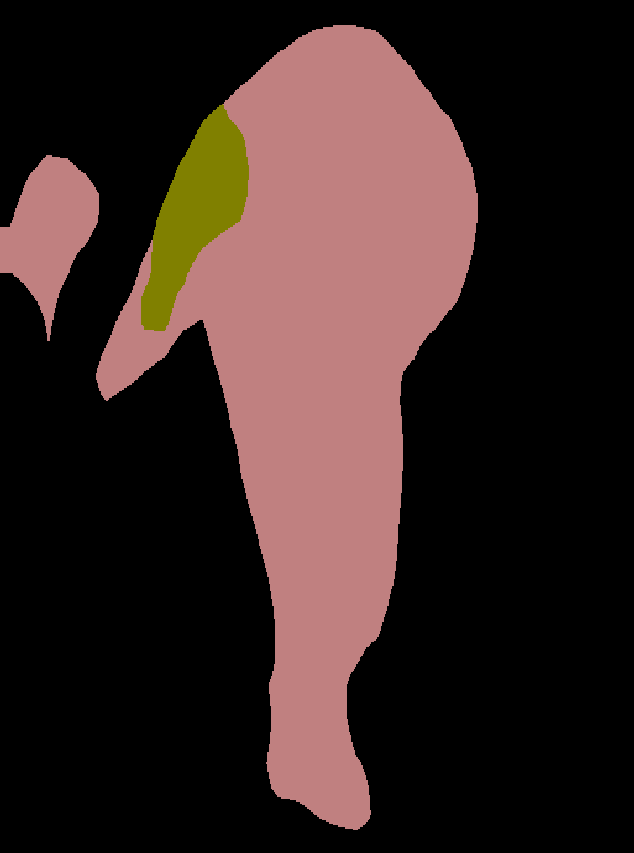}\\

				Image & DeepLabv3~\cite{chen2017rethinking} &Mixup~\cite{zhang2017mixup} & CutMix~\cite{yun2019cutmix} & Binding-CC (ours)& Binding-CM (ours)\\
			\end{tabular}
			
			}
			\caption{\blue{Qualitative examples on the Out-of-Context~\cite{choi2012context} (top five rows) and  UnRel~\cite{peyre2017weakly} (bottom three rows) datasets. Our proposed blending based binding networks (Binding-CC and Binding-CM) generate higher quality segmentation maps compared to the baselines in the out-of-context scenarios.}}
			\label{fig:valcontext}
		\end{center}
\end{figure*}

\subsubsection{Segmenting Highly Occluded Objects in Complex Scenes}\label{sec:occlusion}
We argue that our mixing and source separation strategies are more powerful than existing mixing strategies in complex scenes with large amounts of occlusion. One reason for this is our categorical clustering based mixing strategy (Sec.~\ref{sec:fbt}) blends images based on categorical clusters with dynamic blending ratios. This means that the network will see more images with a wide array of categories blended together, as every category is guaranteed to be blended with every other category. \blue{Additionally, the co-occurrence based mixing strategy blends the images containing semantic objects which are likely to co-occur frequently (e.g., \textit{person} and \textit{motorcycle}). This strategy allows the network to learn stronger representations for objects in commonly occurring complex scenes.} On the other hand, other strategies~\cite{yun2019cutmix,zhang2017mixup} use two randomly selected images to blend. This means the statistics of the generated images will be largely driven by the statistics of the original dataset. Further, the \textit{source separation module }(SSM) is specifically designed for separating features \textit{before} the final layer of the network, allowing for finer details and semantics to be encoded into the target and \textit{phantom} streams. For the other methods, they have a single prediction, which does not allow for these details to be separated early enough in the network to encode as much information as our method.

To substantiate this claim we evaluate each method under three specific data distributions that range in the amount of occlusion and complexity: (i) \textit{Occlusion}: at least one object has occlusion with any other object (1-Occ) in an image and all objects have occlusion (All-Occ), (ii) \textit{Number of Objects}: total number of object instances regardless of classes, and (iii) \textit{Number of Unique Objects}: total number of unique semantic categories. The results are presented in Table ~\ref{table:occlusion}. Our methods outperform the other mixing based methods in all cases. \blue{Interestingly, co-occurrence based blending techniques outperform the clustering based blending under most of the data settings.} Note that the improvements on all occlusion and larger number of unique categories cases are particularly pronounced for our binding models as the performance drop is significantly less than the other methods, when only considering images with many unique objects.

We next perform a cross-dataset experiment by taking our model trained on the PASCAL VOC 2012 training set and evaluate on the publicly available Out-of-Context~\cite{choi2012context} and UnRel~\cite{peyre2017weakly} datasets. Figure~\ref{fig:valcontext} visualizes how the segmentation models trained with only VOC 2012 co-occurring objects performs when objects appear without the context seen in training. \blue{Even with such challenging images with out of context objects (person \textit{on top} of car (see second row)), our methods produce robust segmentation masks while the baselines fail to segment the objects with detail. The co-occurrence based binding network also produces robust segmentation maps despite the nature of training where we blend images with semantic objects which are likely to co-occur. Since Out-of-Context and UnRel datasets do not provide segmentation ground-truth we cannot report quantitative results on these datasets. }

\begin{table*}[t]
	\centering
	\def\arraystretch{1.23}
		\setlength\tabcolsep{3.1pt}

	\resizebox{0.98\textwidth}{!}{
		\begin{tabular}{l|cc c ccc  c c ccc}
			\specialrule{1.2pt}{1pt}{1pt}\

			&  \multicolumn{5}{c}{\textit{Co-occur with \textit{person}}}&& \multicolumn{5}{c}{\textit{Exclusive}} \\
			\cline{2-6} \cline{8-12} 
			& \textbf{horse} & \textbf{mbike} & \textbf{bicycle} & \textbf{bottle} & \textbf{car} & & \textbf{horse} & \textbf{mbike} & \textbf{bicycle} & \textbf{bottle} & \textbf{car} \\
			\hline
			\hspace{0.8cm} \# of Images & 32 & 34 & 30 & 20 & 45 & & 44 & 23 & 29 & 35 & 45 \\
		    \specialrule{1.2pt}{1pt}{1pt}\	
			DeepLabv3-ResNet101~\cite{chen2017rethinking} & 87.9 & 81.6 & 77.7&\textbf{89.7} & \textbf{89.7} && 90.9 & 91.5 & 60.4 & 85.4 & 96.0    \\
			DeepLabv3 + Mixup~\cite{zhang2017mixup} & 86.9 & 82.8 & 76.5 & 87.6 &86.2 && 92.5 & 93.0 & 60.0 & 80.6 & 95.5  \\
			DeepLabv3 + CutMix~\cite{yun2019cutmix} &  86.2 & 83.6 & 76.0 & 87.4 & 87.9 && \textbf{94.1} & \text{93.8} & 61.3 & 82.6 & 96.2\\

			\textbf{DeepLabv3 + Binding (CC)} & \text{89.1} & \textbf{87.2} & \text{78.5} & 86.9 & 89.0&& 94.0 & \text{93.8} & \textbf{61.5} & \text{87.9} & \text{96.4} \\

			
			\textbf{DeepLabv3 + Binding (CM)} & \textbf{89.9} & 86.7 & \textbf{79.3} & 88.5 & 89.2 && 93.6 & \textbf{95.1} & \text{60.2} & \textbf{88.2} & \textbf{96.6}\\

			\specialrule{1.2pt}{1pt}{1pt}
		\end{tabular}}
		\caption{mIoU results on the PASCAL VOC 2012 val set, for the co-occurrence of the most salient person category with five other categories and the results when these five categories appear alone.}
		\label{table:exclusive}
\end{table*}

\subsubsection{Segmenting Out-of-Context Objects}\label{sec:context}
A model that heavily relies on context would not be able to correctly segment compared to a model that truly understands what the object is irrespective of its context. \blue{We argue that our clustering based mixing strategy performs better in out-of-context scenarios, as category-based mixing reduces bias in the dataset's co-occurrence matrix. In contrast, the co-occurrence based blending technique allows the binding network to separate semantic objects which are likely to co-occur. We conduct two experiments to quantitatively evaluate each method's ability to segment out-of-context objects.} 

For the first experiment, we identify the top five categories that frequently co-occur with \textit{person} in the training set, since person has the most occurrences with all other categories based on the co-occurrence matrix. We report performance in Table~\ref{table:exclusive} on two different subsets of data: (i) \textit{Co-occur with Person}: images with both the person and object in it, and (ii) \textit{Exclusive}: images with only the single object of interest. As can be seen from Table~\ref{table:exclusive}, when bottle co-occurs with person all the methods are capable of segmenting bottle and person precisely, whereas the IoU for bottle is significantly reduced when bottle occurs alone. However, our proposed methods (especially the Binding-CC) successfully maintain performance on the exclusive case. 

\begin{figure}[t]
       \begin{tikzpicture}
            \begin{axis}[
        xlabel={Co-occurrence threshold},
        ylabel={mIoU (\%)},
        xmin=2, xmax=65,
        ymin=48, ymax=82,
        xtick={10,20,30,40,50,60},
        xticklabels={10,20,30,40,50,Any},
        x tick label style={font=\footnotesize},
        y tick label style={font=\footnotesize},
        ytick={45,50,60,70,80},
        x dir=reverse,
        width=7.9cm,
        height=5.8cm,
        legend columns=1, 
        legend style={
            cells={anchor=east},
            legend cell align={left},
            draw=none,
            nodes={scale=0.9, transform shape},
            legend style={at={(0.78,0.53)},anchor=south}},
        ymajorgrids=false,
        x label style={at={(axis description cs:0.5,-0.1)},anchor=north,font=\footnotesize},
        y label style={at={(axis description cs:-0.077,.5)},anchor=south,font=\footnotesize},
        grid style=dashed,
    ]
    
    \addplot[line width=1.6pt,mark size=1.3pt,smooth,color=yellow,mark=*,]
        coordinates {(60,77.1)(50,63.7)(40,63.9)(30,63.9)(20,60.1)(10,50.2)};
    
    \addplot[line width=1.6pt,mark size=1.3pt,smooth,color=cyan,mark=*,]
        coordinates {(60,78.9)(50,68.0)(40,65.6)(30,65.5)(20,60.2)(10,53.3)};
        
    \addplot[line width=1.6pt,mark size=1.3pt,smooth,color=orange,mark=*,]
        coordinates {(60,78)(50,66.1)(40,65.2)(30,64.7)(20,61.3)(10,52.9)};
    \addplot[line width=1.6pt,mark size=1.2pt,smooth,color=blue,mark=square*,]
        coordinates {(60,76)(50,60.1)(40,59.9)(30,60.1)(20,56.2)(10,48.8)};
    \addplot[line width=1.6pt,mark size=1.3pt,smooth,color=red,mark=diamond*,]
        coordinates {(60,75.2)(50,66.2)(40,66.2)(30,66.2)(20,61.5)(10,52.8)};
        \legend{DeepLabv3, Binding-CM, Binding-CC, CutMix, Mixup}
        \end{axis}
    \end{tikzpicture}
    \caption{Performance on images with various levels of object co-occurrence. \textit{Binding-CC}, \textit{Binding-CM}, and Mixup~\cite{zhang2017mixup} perform better on subsets of images with unlikely co-occurrences.}
    

    \label{fig:frame}
	
\end{figure}
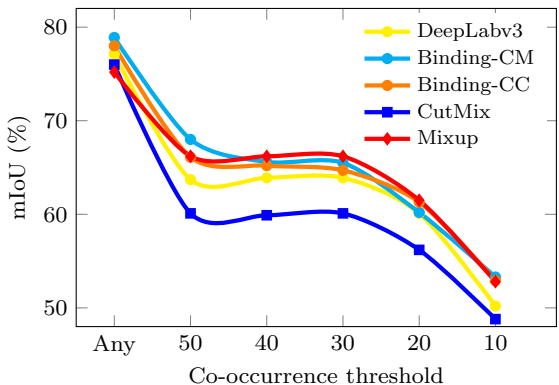

For the second out-of-context experiment, we first create different subsets of images from the VOC 2012 val set based on the training set's co-occurrence matrix. We select thresholds $\{50, 40, 30, 20, 10\}$, and only keep images which have objects that occur less than the chosen threshold. For instance, the threshold value 50 includes all the images where the co-occurrence value of object pairs is less than 50 (e.g., \textit{cat} and \textit{bottle} occur 18 times together, therefore images containing both will be in all subsets except the threshold of 10). Figure~\ref{fig:frame} illustrates the result of different baselines and our methods with respect to co-occurrence threshold. Our methods outperform the baseline DeepLabv3-ResNet101 for all the threshold values. Surprisingly, Mixup~\cite{zhang2017mixup} achieves very competitive performance under few co-occurrence thresholds. \blue{In addition, the co-occurrence based blending network marginally outperforms the clustering based technique which further strengthens the claim that our co-occurrence based blending technique allows the network to better separate semantic objects which are more likely to co-occur.}

\subsection{Adversarial Robustness}\label{sec:adver}
Existing defence mechanisms~\cite{arnab2018robustness,xie2017adversarial,guo2017countering,madry2017towards} against adversarial attacks~\cite{goodfellow2014explaining,kurakin2016adversarial,moosavi2016deepfool,moosavi2017universal} attempt to reduce the impact of adversarial examples. Typically, adversarial defence mechanisms follow two main directions: (i) simply modifying the classifier to make it more robust, or (ii) transforming the adversarial examples in inference time. Even though our pipeline does not fall under either of these categories, 
we further claim our technique works as an implicit defense mechanism against adversarial images similar to~\cite{yun2019cutmix,zhang2017mixup,inoue2018data,cubuk2019autoaugment,harris2020fmix,chou2020remix}. This is because the network optimization, in the form of source separation to solve the binding problem, enhances the capability of interacting with noisy features while imposing a high degree of resilience to interference from the superimposed image.\\
\begin{table}[t]
     \centering
	\def\arraystretch{1.25}
	\setlength\tabcolsep{3.1pt}

	\resizebox{0.49\textwidth}{!}{
		\begin{tabular}{l|c|cc c ccc}
			\specialrule{1.2pt}{1pt}{1pt}\	
			
			\multirow{3}{*}{\hspace{0.1cm} Networks}& \multirow{3}{*}{Clean} & \multicolumn{6}{c}{\textbf{Adversarial Images}}\\
			\cline{3-8}
			& & \multicolumn{2}{c}{UAP~\cite{moosavi2017universal}}&& \multicolumn{3}{c}{GD-UAP~\cite{mopuri2018generalizable}}  \\
			
			\cline{3-4} \cline{6-8}
			
			&&ResNet&GNet& &R-No&R-All &R-Part \\
			
		\specialrule{1.2pt}{1pt}{1pt}\	
			
			DeepLabv3 & 75.9&59.1 & 63.6 & & 66.7 & 63.9& 64.0     \\
			+ Mixup & 75.2 & 62.9& 63.2 && 65.3 & 63.2 & 63.6    \\
			
			+ CutMix & 76.2 & 60.9& 64.3 && 64.2& 62.5& 62.2  \\
			 + \textbf{Binding-CC} & \text{77.9} &\textbf{69.1} & \textbf{70.2}&&  \textbf{68.2}& \textbf{67.0} & \textbf{67.2}   \\	
			 
			 	 
			 	  + \textbf{Binding-CM} &\textbf{ 78.9} & 63.2&67.1 && \text{67.2} & \text{65.0}&\text{64.9}  \\
			
			\specialrule{1.2pt}{1pt}{1pt}
			
		\end{tabular}}
		\caption{Adversarial segmentation robustness performance (mIoU) against the UAP~\cite{moosavi2017universal} and GD-UAP~\cite{mopuri2018generalizable} attacks.}
		\label{table:noisy-analysis}

\end{table}


\begin{figure*} [t]
	\centering
	\setlength\tabcolsep{0.8pt}
	\def\arraystretch{0.5}
	\resizebox{0.98\textwidth}{!}{
		\begin{tabular}{c}
			
			\begin{tabular}{ccc c c cc}

				\includegraphics[width=0.15\textwidth]{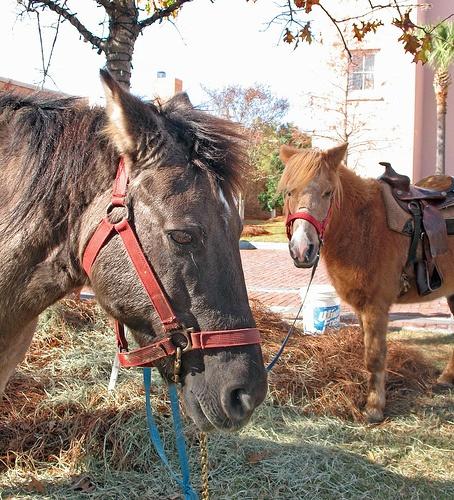}&
				\includegraphics[width=0.15\textwidth]{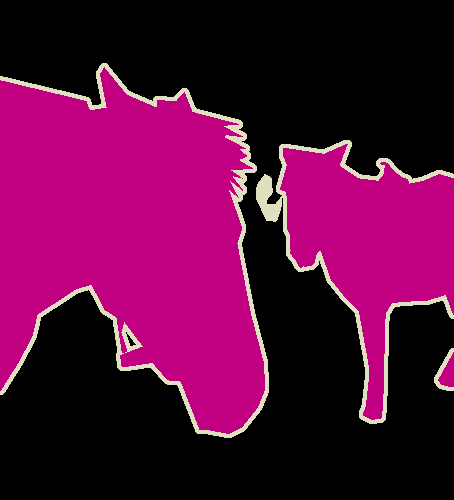}&
				\includegraphics[width=0.15\textwidth]{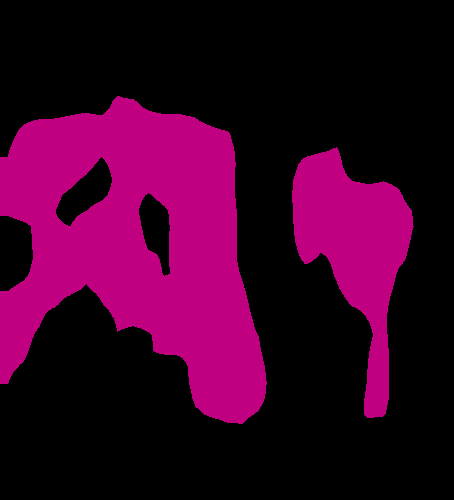}&
				\includegraphics[width=0.15\textwidth]{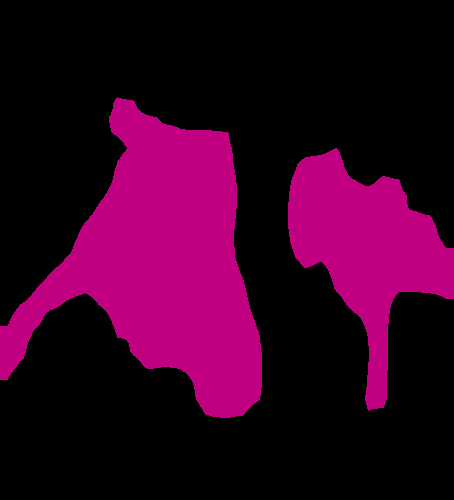}&
				\includegraphics[width=0.15\textwidth]{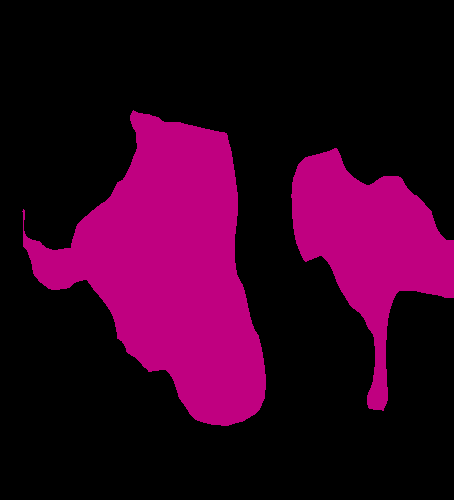}&
	
				\includegraphics[width=0.15\textwidth]{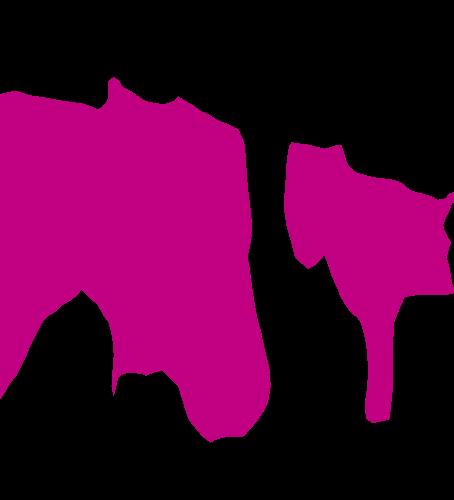}&
				\includegraphics[width=0.15\textwidth]{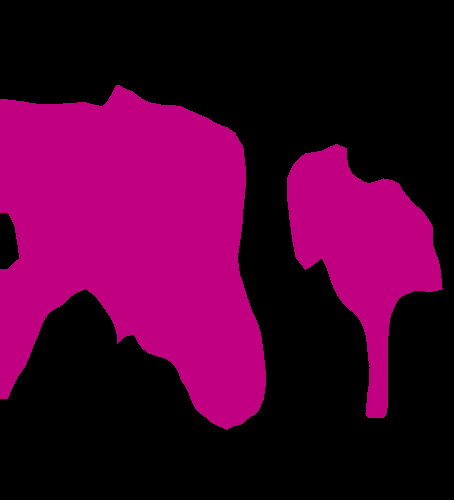}\\
				
		        \includegraphics[width=0.15\textwidth]{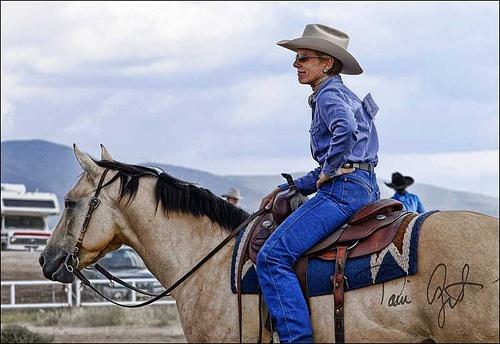}&
				\includegraphics[width=0.15\textwidth]{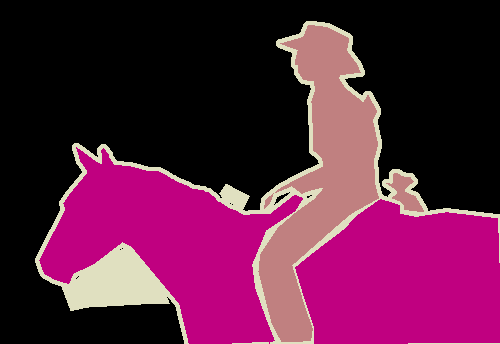}&
				\includegraphics[width=0.15\textwidth]{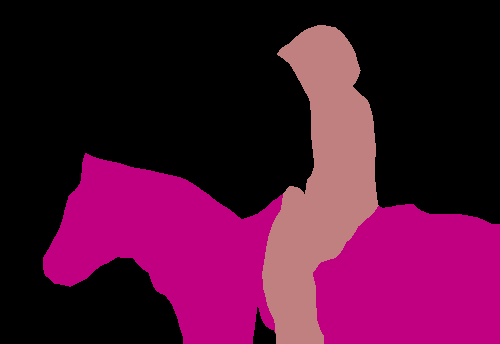}&
				\includegraphics[width=0.15\textwidth]{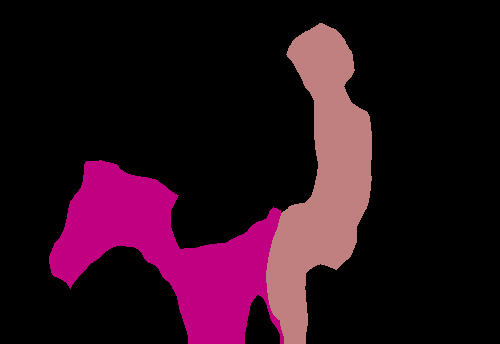}&
				\includegraphics[width=0.15\textwidth]{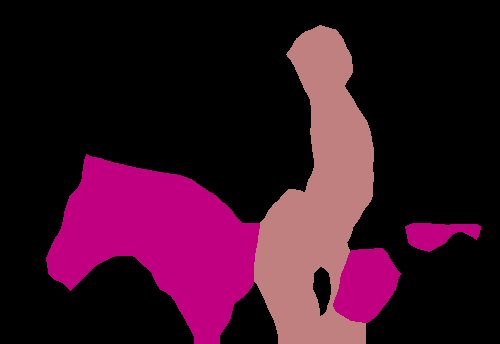}&
	
				\includegraphics[width=0.15\textwidth]{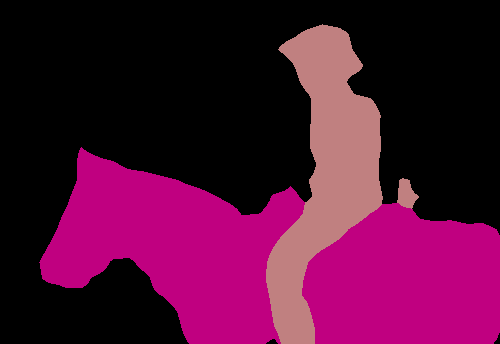}&
				\includegraphics[width=0.15\textwidth]{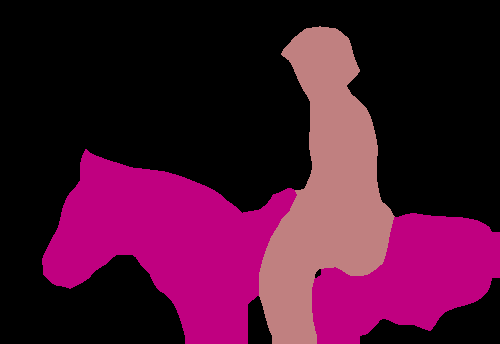}\\
				
				 \includegraphics[width=0.15\textwidth]{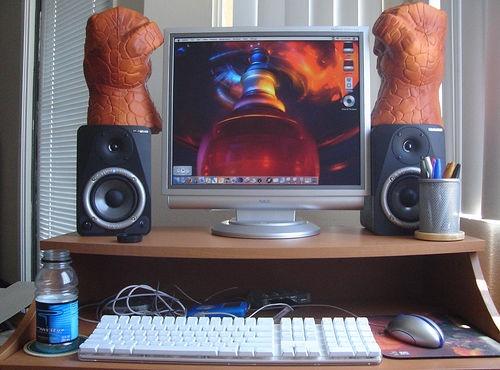}&
				\includegraphics[width=0.15\textwidth]{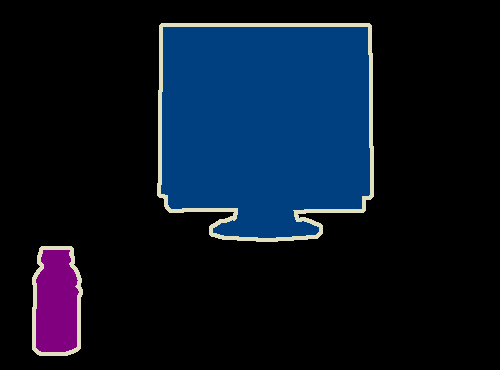}&
				\includegraphics[width=0.15\textwidth]{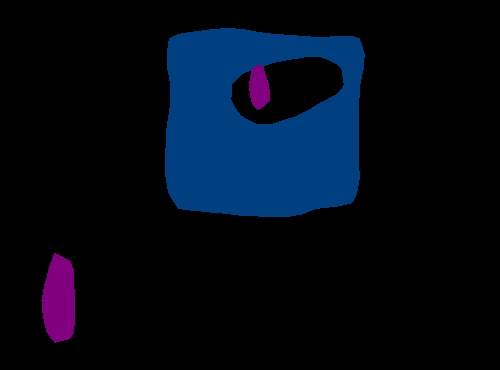}&
				\includegraphics[width=0.15\textwidth]{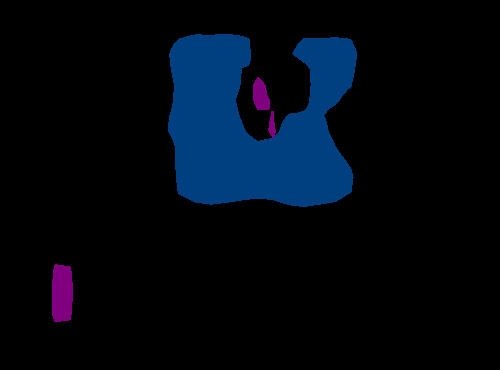}&
				\includegraphics[width=0.15\textwidth]{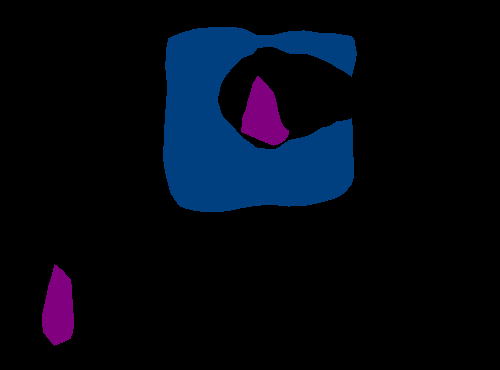}&
	
				\includegraphics[width=0.15\textwidth]{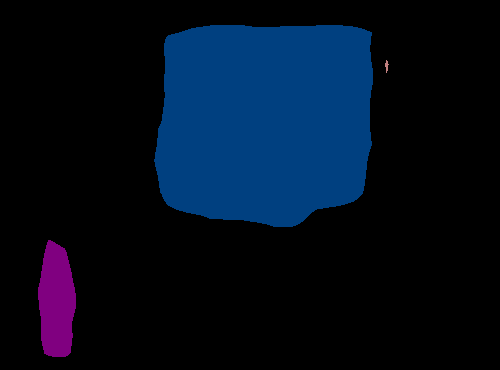}&
				\includegraphics[width=0.15\textwidth]{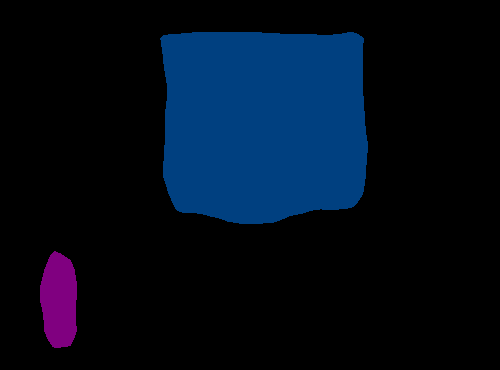}\\

				Image & Ground-truth & DeepLabv3~\cite{chen2017rethinking} &  Mixup~\cite{zhang2017mixup} &  CutMix~\cite{yun2019cutmix} &  Binding-CC (ours) & Binding-CM (ours)
				

			\end{tabular}
			
		\end{tabular}}
		\caption{ Comparison of baselines and our methods when attacked by GD-UAP~\cite{mopuri2018generalizable} under no data settings. Interestingly, the attack is more effective on the baselines compared to the network trained with our methods (Binding-CC and Binding-CM). }
		\label{fig:robust}
		\vspace{-0.2cm}
\end{figure*}

\noindent \textbf{Adversarial Attacks.} We generate adversarial examples using various techniques, including the Universal Adversarial Perturbation (UAP)~\cite{moosavi2017universal} and Generalizable Data-free Universal Adversarial Perturbation (GD-UAP)~\cite{mopuri2018generalizable} under different settings. We use publicly available computed universal perturbations from these methods to generate adversarial examples for the PASCAL VOC 2012 val set. For UAP, which is a black-box attack, we generate adversarial images with both ResNet152 and GoogleNet based universal perturbations. GD-UAP is a grey-box attack, as it generates a perturbation based on the source data (VOC 2012 train set) and the backbone network (ResNet101). For GD-UAP, we compare different levels of adversarial attack strength by generating the perturbation based on various amounts of source data information.\\

\noindent \textbf{Robustness of Segmentation Networks.} We evaluate the robustness of different methods to adversarial examples and show how feature binding-driven training learns to significantly mitigate performance loss due to perturbation. Table~\ref{table:noisy-analysis} shows the robustness of different baselines and our approaches on the PASCAL VOC 2012 validation dataset. In general, DeepLab-based methods~\cite{chen2018deeplab} achieve higher mIoU for the segmentation task on clean examples and are also shown to be more robust to adversarial samples compared to the shallower networks~\cite{arnab2018robustness}. In the case of black-box attacks, the adversarial examples originally generated by UAP on ResNet152, are less malignant when the clustering based blending method is applied in the binding network, while being effective in significantly reducing the performance of other methods.  

When we apply a gray-box attack under the setting (R-All), where VOC 2012 training data and the ResNet101 network are used to generate the perturbation, DeepLabv3 and Mixup show robustness against adversarial examples which is improved by applying our blending strategies. Surprisingly, the performance of CutMix is significantly reduced when tested against adversarial samples generated by GD-UAP. Similarly, we find that DeepLabv3, Mixup, and CutMix are also vulnerable to adversarial cases under the \textit{R-No} and \textit{R-Part} settings, where no data and partial data is used, respectively to generate the perturbations. Notably, DeepLabv3+Binding-CC and DeepLabv3+Binding-CM exhibit significant robustness to extreme cases which further reveals the importance of feature binding training pipeline to successfully relate internal activations corresponding to common sources in the adversarial images. In general, the Binding-CC network shows more robustness than the Binding-CM network under various adversarial settings. The reason behind the greater robustness is that the clustering based technique allows the binding network to be trained on a larger set of noisy mixed data, while the co-occurrence based method allows mixing only between images which have semantic objects that are likely to co-occur. 

\blue{Figure~\ref{fig:robust} depicts the outputs of baselines and our approaches to the GD-UAP attack on the PASCAL VOC 2012 validation set. It is clear that our proposed approaches are more robust against the GD-UAP attack compared to the baseline methods. These observations and results on different 
attacks reveal that the relative ranking of adversarial robustness for the different networks is improved
with the addition of our proposed blending based feature binding training.}

\subsection{Results on Salient Object Detection}\label{sec:SOD}

\begin{table}
	\centering
	     \centering
	\def\arraystretch{1.15}
		\setlength\tabcolsep{8.1pt}
	\resizebox{0.49\textwidth}{!}{
		\begin{tabular}{l|c|c}
			\specialrule{1.2pt}{1pt}{1pt}\
			\multirow{2}{*}{Methods}&  \multicolumn{2}{c}{ECSSD~\cite{yan2013hierarchical}}  \\
			\cline{2-3}
			\multicolumn{1}{c|}{}& $F_\beta$$\uparrow$ & MAE $\downarrow$  \\
			
			\specialrule{1.2pt}{1pt}{1pt}
				
			DeepLabv3-ResNet50~\cite{chen2017rethinking} & 0.906 & 0.045  \\

			DeepLabv3 + Mixup~\cite{zhang2017mixup} & 0.893 & 0.057  \\	
			
			DeepLabv3 + CutMix~\cite{yun2019cutmix} & 0.903 & 0.050  \\	
			
			DeepLabv3 + Binding-CM & \textbf{0.909} & \textbf{0.043}  \\

			\specialrule{1.2pt}{1pt}{1pt}
	\end{tabular}}
	\caption{\blue{Quantitative comparison (in terms of max $F_\beta$ and MAE) with recent methods. Down arrow means lower is better and up arrow means higher is better.}}
	\vspace{-0.3cm}
	\label{table:quant_sal}
\end{table}

\blue{We further validate our proposed co-occurrence based mixing technique on the salient object detection (SOD) task and present a comparison with existing mixing methods~\cite{yun2019cutmix,zhang2017mixup} in Table~\ref{table:quant_sal}. Similar to the task of semantic segmentation, we train the DeepLabv3-ResNet50~\cite{chen2017rethinking} network with various mixing strategies on the DUT-S dataset~\cite{DUTS} and evaluate on ECSSD dataset~\cite{ECSSD}. Since DUT-S dataset does not provide any semantic segmentation ground-truth, we can not directly apply our co-occurrence based image blending technique during training. Towards this goal, we first generate pseudo semantic labels for DUT-S by simply passing the images to the DeepLabv3-ResNet50 network trained on PASCAL VOC 2012 dataset for semantic segmentation task. While the boundaries of the generated pseudo-labels are not perfect, the predicted class labels can still be used as image-level labels in the image blending process.}

\blue{From Table~\ref{table:quant_sal},  it can be seen that our DeepLabv3+Binding-CM method outperforms or achieves competitive performance compared to the baseline methods.} 

\section{Ablation Studies}\label{sec:ablation}
In this section, we examine the variants of our proposed pipelines by considering three different settings: (i) effectiveness of the feature denoising stage and feature binding head (ii) influence of the co-occurrence based mixing technique, and (iii) impact of choosing maximum semantic categories in the co-occurrence based blending. \\

\begin{table}
	\begin{center}
		\def\arraystretch{1.25}
		\setlength\tabcolsep{6.3pt}
		\resizebox{0.48\textwidth}{!}{
			\begin{tabular}{l|c}
				\specialrule{1.2pt}{1pt}{1pt}
				
				\multicolumn{1}{c|}{Methods} & mIoU \\
				\specialrule{1.2pt}{1pt}{1pt}
				DeepLabv3-ResNet50~\cite{chen2017rethinking} & 75.9\\
				DeepLabv3 + Binding (CC) (w/o DN) & 75.4 \\
				DeepLabv3 + Binding (CC)  (w/ DN) & \textbf{75.7} \\

				DeepLabv3 + Binding (CM) (w/o DN) &  76.1\\
				DeepLabv3 + Binding (CM)  (w/ DN) & \textbf{76.2}\\
				
				\midrule
				
				DeepLabv3-ResNet101~\cite{chen2017rethinking} & 77.1 \\
				DeepLabv3 + Binding (CC) (w/o DN) &  76.4\\
				DeepLabv3 + Binding (CC)  (w/ DN) & \textbf{77.9} \\

				DeepLabv3 + Binding (CM) (w/o DN) &  78.3\\
				DeepLabv3 + Binding (CM)  (w/ DN) & \textbf{78.9}\\

				\specialrule{1.2pt}{1pt}{1pt}
			\end{tabular}}

		\caption{\blue{Significance of feature denoising stage (DN). It is clear that the feature denoising stage further improves the overall performance under both mixing techniques.}}
		\label{tab:fb_quan2}
	\end{center}
\end{table}

\begin{table}
	\begin{center}
		\def\arraystretch{1.15}
		\setlength\tabcolsep{3.3pt}
				\resizebox{0.48\textwidth}{!}{
		    \begin{tabular}{l|c}
				\specialrule{1.2pt}{1pt}{1pt}
				
				\multicolumn{1}{c|}{Methods} & mIoU \\
				\hline
				DeepLabv3-ResNet101~\cite{chen2017rethinking} & 77.1 \\
				
				DeepLabv3 + Binding-CC (w/o FBH) & 76.1 \\
				DeepLabv3 + Binding-CC (w/ FBH) & \textbf{76.4} \\

				DeepLabv3 + Binding-CM (w/o FBH) & 77.9 \\
				DeepLabv3 + Binding-CM (w/ FBH) & \textbf{78.3} \\

				\specialrule{1.2pt}{1pt}{1pt}
			\end{tabular}
		}
		\caption{\blue{Performance comparison with and without the feature binding head (FBH) in the source separator module. Including the feature binding head marginally improves the overall performance for both techniques. Note, we report the numbers without the denoising stage.}}
		\label{tab:fb_quan3}
	\end{center}
\end{table}

\subsection{Feature Denoising and Feature Binding Head} We examine the effectiveness of the feature denoising (DN) stage and report results in Table~\ref{tab:fb_quan2}. \blue{Interestingly, using the DeepLabv3-ResNet101~\cite{chen2017rethinking} network as the backbone, the clustering based mixing approach exhibits larger improvement with the addition of the denoising stage than the co-occurrence based technique (1.5\% vs. 0.6\% improvement). The reason behind the larger improvement is that the clustering based technique allows the binding network to be trained on a larger set of noisy mixed data, while the co-occurrence based method allows mixing only between images which have semantic objects that are likely to co-occur. This is why, with a deeper backbone network (e.g., DeepLabv3-ResNet101), the feature binding training with categorical clustering is more noisy which allows the denoising stage to improve the performance more significantly.}

We also conduct experiments (see Table~\ref{tab:fb_quan3}) varying the source separator module, including the feature binding head (FBH). \blue{It is clear that the overall performance of DeepLabv3-ResNet101 based binding networks can be marginally improved with the addition of a feature binding head (0.3\% and 0.4\% improvement respectively)}. We believe the feature binding head allows the network to make a more informed final prediction based on the source \textit{and} the phantom activations, and therefore learns to identify harmful features at inference time, leading to a more accurate prediction. \\

\begin{table}
	\begin{center}
		\def\arraystretch{1.25}
		\setlength\tabcolsep{3.3pt}
				\resizebox{0.49\textwidth}{!}{
		    \begin{tabular}{l|c}
				\specialrule{1.2pt}{1pt}{1pt}
				
				\multicolumn{1}{c|}{Methods} & mIoU \\
					\specialrule{1.2pt}{1pt}{1pt}
				DeepLabv3-ResNet50~\cite{chen2017rethinking} & 75.1 \\
				 + Mixup~\cite{zhang2017mixup} &  73.6\\
				
				 + Mixup~\cite{zhang2017mixup} + Co-occurrence &  74.2\\
				
				 + Mixup~\cite{zhang2017mixup} + Co-occurrence + Binding&  \text{76.1}\\
				 
				 + Mixup~\cite{zhang2017mixup} + Co-occurrence + Binding + DN&  \textbf{76.2}\\
				 
				 \hline
				 
				 DeepLabv3-ResNet101~\cite{chen2017rethinking} &  77.1\\
				 + Mixup~\cite{zhang2017mixup} & 76.2\\
				
				 + Mixup~\cite{zhang2017mixup} + Co-occurrence &  77.2\\
				
				 + Mixup~\cite{zhang2017mixup} + Co-occurrence + Binding&  78.3\\
				 
				 + Mixup~\cite{zhang2017mixup} + Co-occurrence + Binding + DN&  \textbf{78.9}\\

				\specialrule{1.2pt}{1pt}{1pt}
			\end{tabular}
		}
		\caption{\blue{Influence of co-occurrence based image blending techniques and other components on improving overall performance. DN denotes feature denoising stage.}}
		\label{tab:cooccur}
	\end{center}
\end{table}

\subsection{Influence of Co-occurrence based Mixup}
\blue{We further tease out the importance of our proposed co-occurrence based technique by simply applying it with an existing mixup technique. Table~\ref{tab:cooccur} presents quantitative results comparing different components. DeepLabv3-ResNet50 with Mixup~\cite{zhang2017mixup} achieves 73.6\% mIoU. The performance is improved by 0.6\% when we apply co-occurrence matrix based blending with Mixup. The feature binding training pipeline further improves the overall performance by 1.9\% which is further improved by 0.1\% by applying the denoising stage. From the results, it is clear that co-occurrence based blending has an clear influence on improving the segmentation performance. As shown in Table~\ref{tab:cooccur}, the results are consistent when we use DeepLabv3-ResNet101 as the backbone.}

\begin{table}
	\begin{center}
		\def\arraystretch{1.2}
		\setlength\tabcolsep{3.3pt}
		\resizebox{0.48\textwidth}{!}{
			\begin{tabular}{l|c}
				\specialrule{1.2pt}{1pt}{1pt}
				
				\multicolumn{1}{c|}{Methods} & mIoU \\
				\specialrule{1.2pt}{1pt}{1pt}
				
				DeepLabv3-ResNet50~\cite{chen2017rethinking} & 75.1 \\
			
				DeepLabv3 + Binding-CM (max = 2) &  \text{75.9}\\
				DeepLabv3 + Binding-CM (max = 3)   & \textbf{76.1}\\
				DeepLabv3 + Binding-CM (max = 4)   & 75.6\\
				DeepLabv3 + Binding-CM (max = $\infinity$) &  \text{76.1}\\
				DeepLabv3 + Binding-CM (max = $\infinity$) + $\gamma$-thres &  \textbf{76.1}\\
				
			\midrule	DeepLabv3-ResNet101\cite{chen2017rethinking} & 77.1 \\
			
				DeepLabv3 + Binding-CM (max = 2) &  \text{77.2}\\
				DeepLabv3 + Binding-CM (max = 3)   & \text{77.6}\\
				DeepLabv3 + Binding-CM (max = 4)   & 77.7\\
				DeepLabv3 + Binding-CM (max = $\infinity$) &  \text{77.0}\\
				
				DeepLabv3 + Binding-CM  (max = $\infinity$) + $\gamma$-thres  &  \textbf{78.3}\\

				\specialrule{1.2pt}{1pt}{1pt}
			\end{tabular}}

		\caption{\blue{We examine the influence of changing the maximum number of unique semantic objects during the blending process. Note, we report the number without any denoising stage. `$\gamma$-thres' refers to the default model which allows the blended images over the maximum threshold but sets the mixing ratio, $\gamma$,  to 0.9.}}
		\label{tab:fb_max_cat}
	\end{center}
\end{table}

\subsection{Impact of Choosing Maximum Semantic Categories in Co-occurrence based Mixup}
\blue{Existing mixing based techniques have been applied mostly on datasets where there exists one dominant semantic object (e.g., ImageNet~\cite{russakovsky2015imagenet}, CIFAR-10). However, semantic segmentation datasets naturally contain images with more than one category in complex scenes. Therefore, randomly combining two source images based on the co-occurrence likelihood during training to achieve the desired objective is still a more significant challenge than one might expect in the context of dense labeling. For instance, if we blend two images containing three and four semantic objects with a lower mixing ratio (i.e., assign more weight to the \textit{target} image), there is a high chance that the mixed image will lack context. Also, images with seven unique objects are extremely rare or non-existent in the datasets we explore, and therefore this type of image may be too different from the target distribution. To explore this issue, we first restrict the number of unique semantic categories to be blended (e.g., we do not blend images if the maximum threshold is exceeded). We additionally try a strategy where we mix the images, but set the mixing ratio to a constant value (0.9) if this threshold is surpassed. The intuition is that, for a pair of images where the total number of unique semantic categories is higher than the threshold, we want to reduce the amount of blending by assigning more weight to the source image. }

Table~\ref{tab:fb_max_cat} presents the results of choosing different thresholds in the co-occurrence based mixing process. It is clear that restricting the maximum number of unique objects by a threshold along with restricting the mixing ratio achieves higher mIoU compared to other alternatives. 

\section{Discussion and Conclusion}\label{sec:conclude}

\begin{figure}
  \centering

	  \begin{center}
		\setlength\tabcolsep{0.3pt}
		\def\arraystretch{0.3}
		\resizebox{0.49\textwidth}{!}{
			\begin{tabular}{*{5}{c}}		

				\includegraphics[width=0.11\textwidth]{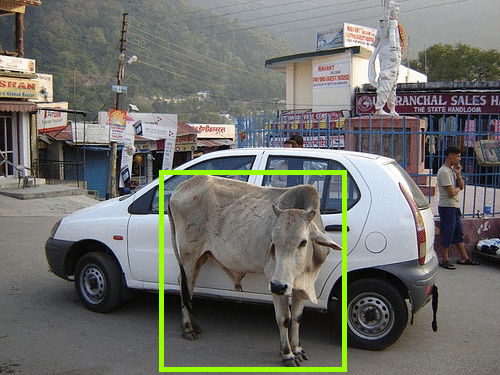}&
				\includegraphics[width=0.11\textwidth]{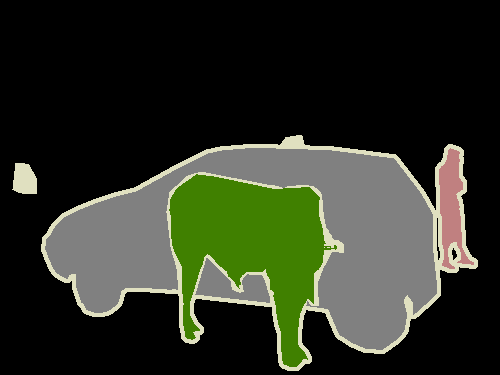}&
				\includegraphics[width=0.11\textwidth]{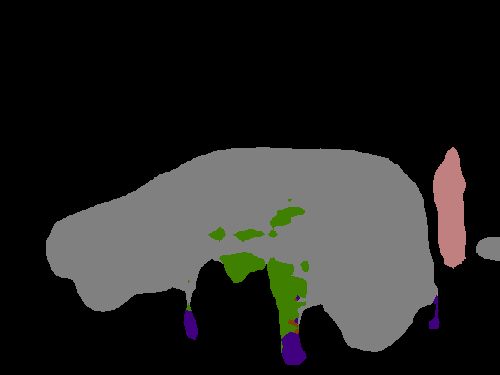}&
				\includegraphics[width=0.11\textwidth]{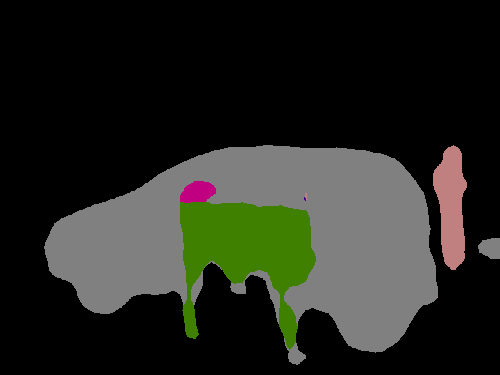}&
				\includegraphics[width=0.11\textwidth]{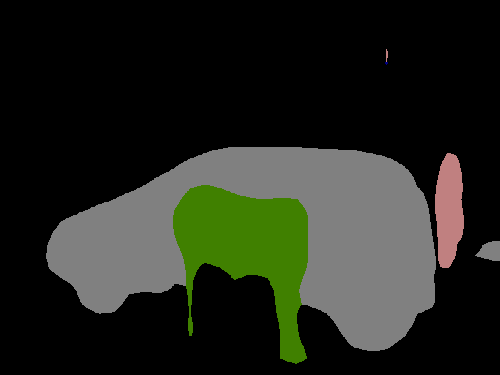}\\		
				
				\includegraphics[width=0.11\textwidth]{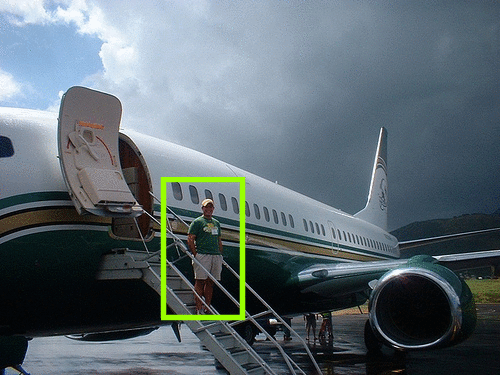}&
				\includegraphics[width=0.11\textwidth]{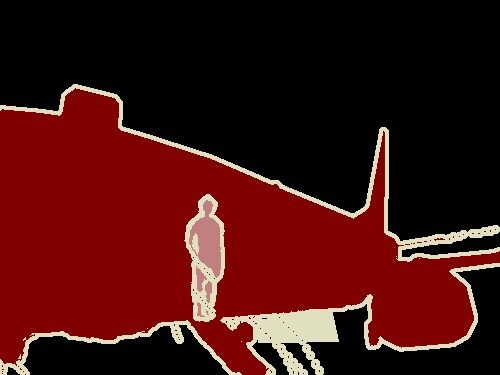}&
				\includegraphics[width=0.11\textwidth]{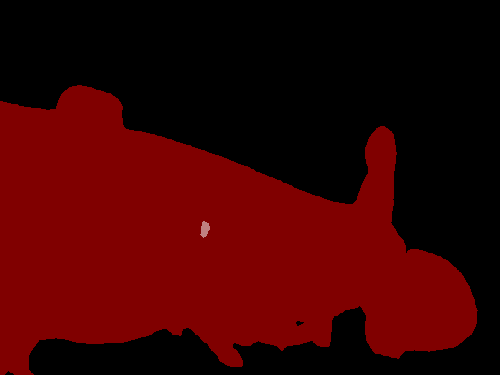}&
				\includegraphics[width=0.11\textwidth]{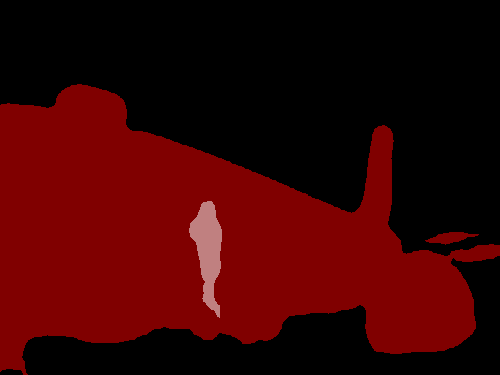}&
				\includegraphics[width=0.11\textwidth]{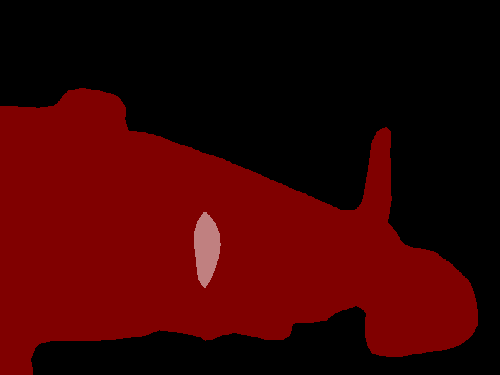}\\

				Image & GT& DeepLabv3 & Binding-CC & Binding-CM
			\end{tabular}
			}
			\vspace{0.2cm}
			\captionof{figure}{Two challenging images where semantic objects are highly occluded. While pixels belonging to the dominant semantic categories are identified correctly, the prediction fails to relate activation tied to smaller occluding features to correct categorical assignments. This is resolved when trained using our proposed mixing based binding networks.}
			\label{fig:discuss}
		\end{center}

\end{figure}
Training with the categorical clustering and a co-occurrence based feature binding pipeline enables learning resilient features, separating sources of activation, and resolving ambiguity with richer contextual information. Although DeepLabv3 is a powerful segmentation network, there are cases (see Fig.~\ref{fig:discuss}) where background objects are correctly classified (car and plane) but other semantic categories are not separated correctly due to high degrees of occlusion (person on the stairs, see Fig.~\ref{fig:discuss} right). In contrast, the feature binding based learning approaches are highly capable of resolving such cases by learning to separate source objects and tying them to specific regions.\\

In summary, we have presented two approaches to train CNNs based on the notion of feature binding. This process includes, as one major component, careful creation of categorical collisions in data during training. This results in improved segmentation performance, and also promotes significant robustness to adversarial perturbations. Denoising in the form of fine-tuning shows further improvement along both these dimensions. 



\clearpage

%
\bibliographystyle{IEEEtran}
\bibliography{paper3,paper}


\end{document}